\documentclass[10pt,journal,compsoc]{IEEEtran}

\usepackage{times}
\usepackage{helvet}
\usepackage{courier}

\usepackage{graphicx}
\usepackage{amsmath,amssymb} 
\usepackage{color}
\usepackage{times}
\usepackage{epsfig}
\usepackage{float}
\usepackage{url}
\usepackage{subfig}
\usepackage{multirow}
\usepackage{comment}

\usepackage{enumitem}

\usepackage{array}
\newcolumntype{x}[1]{>{\centering\arraybackslash\hspace{0pt}}m{#1}}
\usepackage{epstopdf}
\def\footnoterule{\relax%
	\kern-5pt
	\hbox to \columnwidth{\hfill\vrule width \columnwidth height 0.4pt\hfill}
	\kern4.6pt
	}
\makeatother

\newcommand{\cc}{\textcolor{black}}

\newcommand{\KG}{\textcolor{black}}
\newcommand{\KGtwo}{\textcolor{black}}

\newcommand{\iccvf}{\textcolor{black}}
\newcommand{\CR}{\textcolor{black}}
\newcommand{\BX}{\textcolor{black}}
\newcommand{\BO}{\textcolor{black}}

\newcommand{\bx}{\textcolor{black}}
\newcommand{\bbx}{\textcolor{black}}

\ifCLASSOPTIONcompsoc
  \usepackage[nocompress]{cite}
\else
  \usepackage{cite}
\fi
\ifCLASSINFOpdf
\else
\fi
\begin{document}

%
\title{Pixel Objectness: Learning to Segment Generic Objects Automatically in Images and Videos}
%
%
%
%

\author{Bo~Xiong\thanks{\IEEEauthorrefmark{1} Both authors contributed equally to this work}\IEEEauthorrefmark{1},
Suyog~Dutt~Jain\IEEEauthorrefmark{1}, and Kristen~Grauman,~\IEEEmembership{Member,~IEEE}
\IEEEcompsocitemizethanks{\IEEEcompsocthanksitem Bo Xiong, Suyog Dutt Jain, and Kristen Grauman
are with the Department of Computer Science, University of Texas at Austin.\protect\\

}
}
%
%

\markboth{IEEE TRANSACTIONS ON PATTERN ANALYSIS AND MACHINE INTELLIGENCE}%
{Shell \MakeLowercase{\textit{et al.}}: IEEE TRANSACTIONS ON PATTERN ANALYSIS AND MACHINE INTELLIGENCE}

\IEEEtitleabstractindextext{%
\begin{abstract}
We propose an end-to-end learning framework for segmenting generic objects in both images and videos. Given a novel image or video, our approach produces a pixel-level mask for all ``object-like" regions---even for object categories never seen during training. We formulate the task as a structured prediction problem of assigning an object/background label to each pixel, implemented using a deep fully convolutional network.  \KG{When applied to a video, our model further incorporates a motion stream, and the network learns to combine both appearance and motion and \KGtwo{attempts to extract} all prominent objects whether they are moving or not.    
Beyond the core model, a second contribution of our approach is how it leverages varying strengths of training annotations.
Pixel-level annotations are quite difficult to obtain, yet crucial for training a deep network approach for segmentation.  Thus we propose ways to exploit weakly labeled data for learning dense foreground segmentation.  For images,} we show the value in mixing object category examples with  \emph{image-level} labels together with relatively few images with \emph{boundary-level} annotations.  \KG{For video,} we 
 show how to bootstrap weakly annotated videos together with the network trained for image segmentation. Through experiments on multiple challenging image and video segmentation benchmarks, \KG{our method offers consistently strong results} and improves the state-of-the-art for fully automatic segmentation of generic (unseen) objects. 
 In addition, we demonstrate how our approach benefits image retrieval and image retargeting, both of which flourish when given our high-quality foreground maps.  \KGtwo{Code, models, and videos are at: \url{http://vision.cs.utexas.edu/projects/pixelobjectness/}}

\end{abstract}

}

\maketitle

\IEEEdisplaynontitleabstractindextext

%
\IEEEpeerreviewmaketitle

\section{Introduction}\label{sec:introduction}

\begin{figure}[t]
\centering
\includegraphics[width=1\columnwidth]{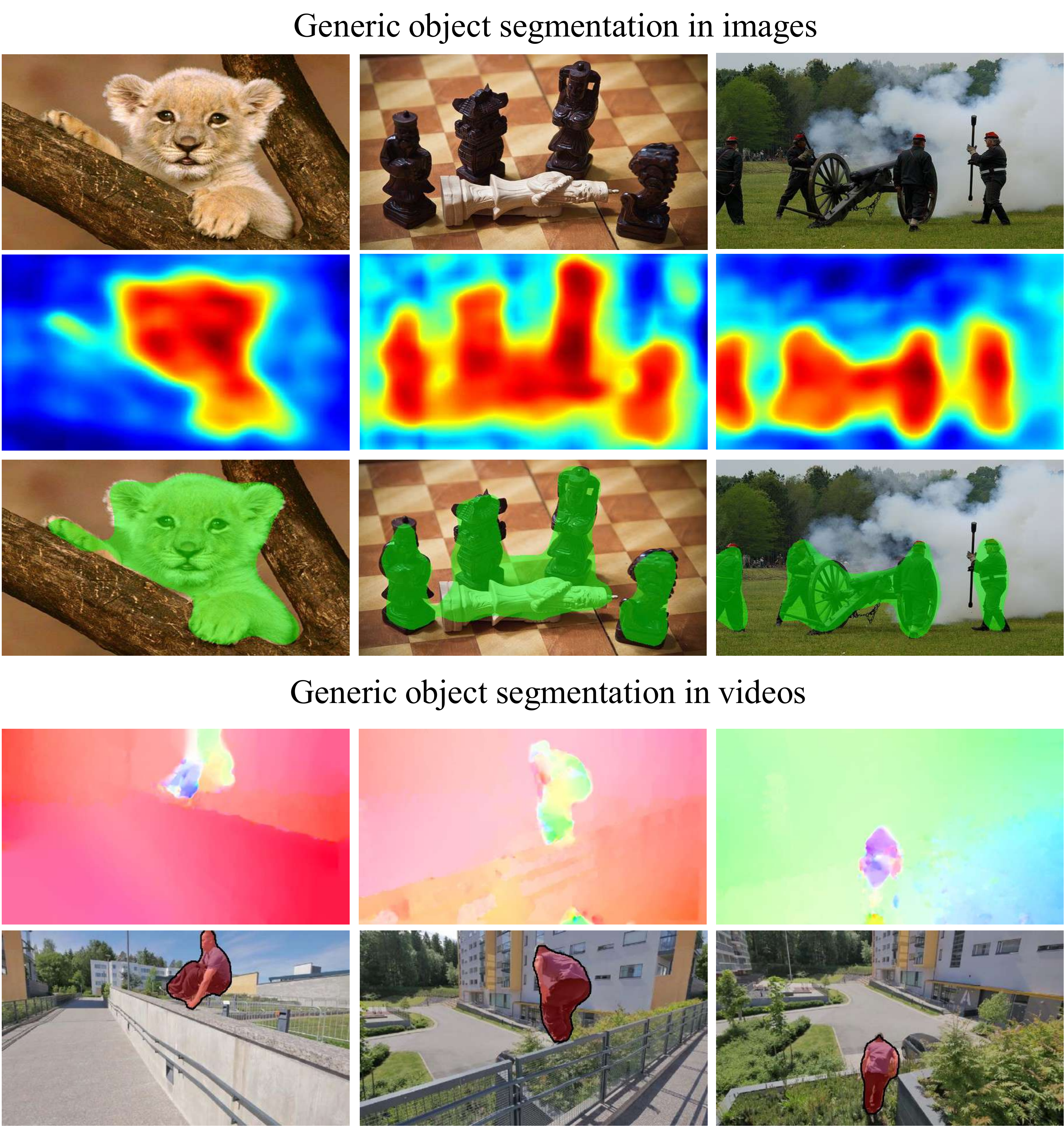}
\caption{Given a novel image (top row), our method predicts an objectness map for each pixel (2nd row) and a single foreground segmentation (3rd row).  Given a novel video, our end-to-end trainable model simultaneously draws on the strengths of generic object appearance and motion (4th row, color-coded optical flow images) to extract generic objects (last row).}
\label{fig:concept}
\vspace{8pt}
\end{figure}

\IEEEPARstart{G}{eneric} \KGtwo{foreground} object segmentation in images and videos is a fundamental vision problem with several applications. For example, a visual search system can use generic object segmentation to focus on the important objects in the query image, ignoring background clutter.  It is also a prerequisite in graphics applications like image retargeting, production video editing, and rotoscoping.  Knowing the spatial extent of objects can also benefit downstream vision tasks like scene understanding, caption generation, and summarization. In any such setting, it is crucial to segment ``generic" objects in a \emph{category-independent} manner.  That is, the system must be able to identify object boundaries for objects it has never encountered during training.  This differentiates the problem from traditional recognition or ``semantic segmentation"~\cite{long_shelhamer_fcn,chen14semantic}, where the system is trained specifically for predefined categories, and is not equipped to segment any others.

Today there are two main strategies for generic object segmentation in images: saliency and object proposals. Saliency methods yield either highly localized attention maps~\cite{LiuHZWL15,pan2016shallow,borji_survey} or a complete segmentation of the prominent object~\cite{zhang2013saliency, czmhh_contrastSaliency_cvpr11,jiangsaliency,liu-salient,secret,zhao2015saliency,DeepSaliency}. Saliency focuses on regions that stand out, which is not the case for all foreground objects. Alternatively, \emph{object proposal} methods learn to localize all objects in an image, regardless of their category~\cite{cpmc,APBMM2014,endres,ZitnickECCV14edgeBoxes,UijlingsIJCV2013,deepmask,Hosang2015Pami}. The aim is to obtain high recall at the cost of low precision, i.e., they must generate a large number of proposals (typically 1000s) to cover all objects in an image. This usually involves a multi-stage process: first bottom-up segments are extracted, then they are scored by their degree of ``objectness".  Relying on bottom-up segments can be limiting, since low-level cues may fail to pull out contiguous regions for complex objects.  Furthermore, in practice, the accompanying scores are not so reliable such that one can rely exclusively on the top few proposals.  

\KG{In video object segmentation, motion offers important additional cues for isolating foreground objects that may be difficult to find in an individual image.  Yet existing methods fall short of leveraging both appearance and motion in a unified manner.  On the one hand, interactive techniques \emph{strongly rely on appearance} information stemming from human-drawn outlines on frames in the video, using motion primarily to propagate information or enforce temporal consistency~\cite{sudheendra-eccv2012,suyog-eccv2014,Perazzi_2015_ICCV}. On the other hand, fully automatic methods \emph{strongly rely on motion} to seed the segmentation process by locating possible moving objects.  Once a moving object is detected, appearance is primarily used to track it across frames~\cite{keysegments,ferrari-iccv2013,nlc}.  Such methods can fail if the object(s) are static or when there is significant camera motion.  In either paradigm, results can suffer because the two essential cues are treated only in a sequential or disconnected way.}

Motivated by these shortcomings, we introduce \emph{pixel objectness}, a new approach to generic \KGtwo{foreground} object segmentation in images and video.\footnote{\KGtwo{An initial version of this work appears in CVPR 2017~\cite{fusionseg,pixelObjectness}}.}  Given a novel image \KG{or video frame}, the goal is to determine the likelihood that each pixel is part of a foreground object (as opposed to background or ``stuff" classes like grass, sky, sidewalks, etc.)  Our definition of a generic object follows that commonly used in the object proposal literature~\cite{Alexe,cpmc,APBMM2014,endres,ZitnickECCV14edgeBoxes,UijlingsIJCV2013}.  Pixel objectness quantifies how likely a pixel belongs to an object of \emph{any} class, and should be high even for objects unseen during training. See Fig.~\ref{fig:concept} (top).  \KG{For the image case, pixel objectness can be seen as a pixel-level extension of window-level objectness~\cite{Alexe}, and hence the name for our method is a nod to that influential work.}

We propose an end-to-end trainable model that draws on the respective strengths of generic (non-category-specific) object appearance and motion in a unified framework. 
Specifically, we develop a novel two-stream fully convolutional deep segmentation network where individual streams encode generic appearance and motion cues derived from a video frame and its corresponding optical flow. These individual cues are fused in the network to produce a final object versus background pixel-level binary segmentation for each video frame (or image). See Fig.~\ref{fig:concept}.  The proposed network segments both static and moving objects without any human involvement.

Declaring that motion should assist in video segmentation is non-controversial, and indeed we are certainly not the first to inject motion into video segmentation.  However, the empirical impact of motion for today's most challenging datasets has been arguably underwhelming. When it comes to motion and appearance,
thus far the sum is not much greater than its parts.  We contend that this is because \emph{the signal from motion is adequately complex such that rich learned models are necessary to exploit it}.  For example, a single object may display multiple motions simultaneously, background and camera motion can intermingle, and even small-magnitude motions should be informative.  \KG{Our approach is capable of learning these factors from data.}

\KG{Three factors suggest that the path we are pursuing may require extensive foreground-annotated examples across a vast array of categories: 1) we wish to handle arbitrary objects of any category, 2) we aim to extract dense pixel-level segment masks, and 3) we train a deep convolutional network for the task.  Unfortunately, dense pixel-level annotations are not easy to come by, particularly for video data.  Hence a second key contribution of our work is to explore how weaker annotations can be adopted to train the models.  First, we show that,} somewhat surprisingly, when training the appearance stream of our model with \emph{explicit boundary-level} annotations for few categories pooled together into a single generic ``object-like" class, pixel objectness generalizes well to \emph{thousands} of unseen objects.  This generalization ability is facilitated by an \emph{implicit image-level} notion of objectness built into a pretrained classification network, which we transfer to our segmentation model during initialization.  \KG{Second, to allow training with few densely labeled video examples,} we show how to leverage readily available \emph{image} segmentation annotations together with \emph{weakly annotated video} data \KG{to train the motion stream of our model.}

Through extensive experiments, we show that our model generalizes very well to unseen objects.  For images, we obtain state-of-the-art performance on the challenging ImageNet~\cite{imagenet_cvpr09} and MIT Object Discovery~\cite{rubinstein-cvpr2013} datasets. We also show how to leverage our segmentations to benefit object-centric image retrieval and content-aware image resizing. For video segmentation, we advance the state-of-the-art for fully automatic video object segmentation on multiple challenging datasets, DAVIS~\cite{Perazzi2016}, YouTube-Objects~\cite{prest2012learning,suyog-eccv2014,tang-cvpr2013}, and Segtrack-v2~\cite{rehg-iccv2013}. Our results show the reward of learning from both signals in a unified framework: a true synergy, often with substantially stronger results than what we can obtain from either one alone---even if they are treated with an equally sophisticated deep network. In summary, we make the following novel contributions:

\begin{itemize}
\item \iccvf{We are the first to show how to train a state-of-the-art generic object segmentation model without requiring a large number of annotated segmentations from thousands of diverse object categories.}

	\item \iccvf{Through extensive results on 3,600$+$ categories and $\sim$1M images, our model generalizes to segment thousands of unseen categories. No other prior work---including recent deep saliency and object proposal methods---shows this level of generalization.}

		\item \KG{Generalizing to video, we also} propose the first end-to-end \CR{trainable} framework for producing pixel level generic object segmentation automatically in videos.  	
		\item Our framework achieves state-of-the-art accuracy on multiple video segmentation datasets, improving over many reported results in the literature and strongly outperforming simpler applications of optical flow. 
		\item We present a means to train a deep pixel-level video segmentation model with access to only weakly labeled videos and strongly labeled images, \KG{with no explicit assumptions about the categories present in either}.
\end{itemize}

\section{Related Work}

We summarize how our ideas relate to prior work in image and video segmentation. For image segmentation, we divide related work into two top-level groups: 1) methods that extract an object mask no matter the object category, and 2) methods that learn from category-labeled data, and seek to recognize/segment those particular categories in new images. Our image segmentation method fits in the first group.  For video segmentation, we summarize fully automatic methods, human-guided methods and deep learning with motion. Our method is fully automatic.

\subsection{Category-independent image segmentation}

{\bf Interactive image segmentation} algorithms such as GrabCut~\cite{grabcut} let a human guide the algorithm using bounding boxes or scribbles. These methods are most suitable when high precision segmentations are required such that some guidance from humans is worthwhile. While some methods try to minimize human involvement~\cite{suyog-iccv2013,pull-plug}, still typically a human is in the loop to guide the algorithm. In contrast, our model is fully automatic and segments objects without any human guidance.

{\bf Object proposal methods}, also discussed above, produce thousands of object proposals either in the form of bounding boxes~\cite{endres,ZitnickECCV14edgeBoxes,UijlingsIJCV2013} or regions~\cite{cpmc,APBMM2014,deepmask,Hosang2015Pami}. Generating thousands of hypotheses ensures high recall, but often results in low precision.  Though effective for object detection, it is difficult to automatically filter out inaccurate proposals without class-specific knowledge. We instead generate a \emph{single} hypothesis of the foreground as our final segmentation.  Our experiments directly evaluate our method's advantage.

{\bf Saliency models} have also been widely studied in the literature. The goal is to identify regions that are likely to capture human attention. While some methods produce highly localized regions~\cite{LiuHZWL15,pan2016shallow,borji_survey}, others segment complete objects~\cite{czmhh_contrastSaliency_cvpr11,jiangsaliency,liu-salient,secret,zhao2015saliency,DeepSaliency}.  While saliency focuses on objects that ``stand out'',  our method is designed to segment all foreground objects, irrespective of whether they stand out in terms of low-level saliency. This is true even for the deep learning based saliency methods~\cite{pan2016shallow,LiuHZWL15,zhao2015saliency,DeepSaliency} which are also end-to-end trained but prioritize objects that stand out.

\subsection{Category-specific image segmentation}

{\bf Semantic segmentation} refers to the task of jointly \emph{recognizing} and segmenting objects, classifying each pixel into one of $k$ fixed categories. Recent advances in deep learning have fostered increased attention to this task. Most deep semantic segmentation models include fully convolutional networks that apply successive convolutions and pooling layers followed by upsampling or deconvolution operations in the end to produce pixel-wise segmentation maps~\cite{long_shelhamer_fcn,chen14semantic}. However, these methods are trained for a fixed number of categories. We are the first to  show that a fully convolutional network can be trained to accurately segment \emph{arbitrary} foreground objects.  Though relatively few categories are seen \KGtwo{with segmentations} in training, our model generalizes very well to unseen categories (as we demonstrate for 3,624 classes from ImageNet, only a fraction of which overlap with PASCAL, the source of our training masks).

{\bf Weakly supervised joint segmentation} methods use weaker supervision than semantic segmentation methods. Given a batch of images known to contain the same object category, they segment the object in each one. The idea is to exploit the similarities within the collection to discover the common foreground. The output is either a pixel-level mask~\cite{vicente-cvpr2011,joulin-cvpr2012,kim-iccv2011,rubinstein-cvpr2013,chen_cvpr14,suyog-cvpr2016} or bounding box~\cite{ferrari_ijcv12,TangCVPR14}.  While joint segmentation is useful, its performance is limited by the shared structure within the collection; intra-class viewpoint and shape variations pose a significant challenge. Moreover, in most practical scenarios, such weak supervision is not available.  A stand alone single-image segmentation model like ours is more widely applicable.

{\bf Propagation methods} transfer information from exemplars with human-labeled foreground masks~\cite{Kuettel2012cvpr,guillaumin2014imagenet,suyog-cvpr2016}. They usually involve a matching stage between likely foreground regions and the exemplars. The downside is the need to store a large amount of exemplars at test time and perform an expensive and potentially noisy matching process for each test image. In contrast, our segmentation model, once trained end-to-end, is very efficient to apply and does not need to retain any training data.

\subsection{Automatic video segmentation methods}

\KG{Similar to image segmentation work, video segmentation is explored under varying degrees of supervision.}  Fully automatic or unsupervised video segmentation methods assume no human input. However, while image segmentation relies on appearance cues, video segmentation can also utilize motion. Methods can be grouped into two broad categories. First we have the supervoxel methods~\cite{grundmann-cvpr2010,xu-eccv2012,galasso-accv2012} which oversegment the video into space-time blobs with cohesive appearance and motion. Their goal is to generate mid-level video regions for downstream processing \bx{(e.g., action detection~\cite{yang2017spatio,escorcia2016daps})}, whereas ours is to produce space-time tubes which accurately delineate object boundaries.  

Second we have the fully automatic methods that generate thousands of ``object-like" space-time segments~\cite{Wu_2015_CVPR,Fragkiadaki_2015_CVPR,oneata,xiao-cvpr2016}.
While useful to accelerate object detection, it is not straightforward to automatically select the most accurate one when a single hypothesis is desired.  Methods that do produce a single hypothesis\bx{~\cite{keysegments,ferrari-iccv2013,nlc,Tsai_ECCV_2016,Sundberg,andrew_stein,bideau2016s,koh2017primary,memory}} strongly rely on motion to identify the objects, either by seeding appearance models with moving regions or directly reasoning about occlusion boundaries using optical flow. This limits their ability to segment static objects in video.  In comparison, our method is fully automatic, produces a single hypothesis, and can segment both static and moving objects.  
Concurrent work~\cite{tokmakov2016learning} trains a deep network with synthetic data to predict moving objects from motion. Our works differs in two ways: 1) we show how to bootstrap weakly annotated real videos together with existing image recognition datasets for training; 2) our framework learns from appearance and motion jointly whereas theirs is trained with only motion.

\subsection{Human-guided video segmentation methods}

\KG{Related to the interactive image methods discussed above are approaches for semi-automatic video segmentation.} Semi-supervised label propagation methods accept human input on a subset of frames, then propagate it to the remaining frames~\cite{ren-cvpr2007,cipolla-cvpr2010,fathi-bmvc2011,sudheendra-eccv2012,suyog-eccv2014,Perazzi_2015_ICCV,marki2016bilateral,Tsai_CVPR_2016,caelles2017one,jampani2017video,khoreva2017learning}. In a similar vein, interactive video segmentation methods leverage a human in the loop to provide guidance or correct errors, e.g.,~\cite{bai-2009,Nagaraja_2015_ICCV,price-iccv2009}. Deep learning human-guided video segmentation methods~\cite{caelles2017one,jampani2017video,khoreva2017learning} typically focus on learning object appearance from the manual annotations since the human pinpoints the object of interest. Motion is primarily used to propagate information or enforce temporal smoothness. In the proposed method, both motion and appearance play an equally important role, and we show their synergistic combination results in much better segmentation quality. Moreover, our method is fully automatic and uses no human involvement \CR{to segment a novel video.}

\subsection{Deep learning with motion}
Deep learning for combining motion and appearance in videos has proven to be useful in several other computer vision tasks such as video classification~\cite{BeyondShort,KarpathyCVPR14}, action recognition~\cite{Simonyan,3DAction}, object tracking~\cite{Wang_2015_ICCV,Ma-ICCV-2015} and even computation of optical flow~\cite{Dosovitskiy_2015_ICCV}. While we take inspiration from these works, we are the first to present a deep framework for segmenting objects in videos in a fully automatic manner.

This manuscript builds on our conference paper on video segmentation~\cite{fusionseg}.  The main new additions are 1) all new results on three image datasets demonstrating our approach applied to image segmentation, with comparisons to 16 additional methods in the literature; 2) new applications showing the impact of our image segmentation for image retargeting and retrieval; and 3) an upgrade to the fusion model that results in improved performance.


\section{Approach}
\label{sec:approach1}

Our goal is to predict the likelihood of each pixel being a generic object as opposed to background. 
As defined in the influential work of~\cite{Alexe}, a ``generic object'' should have at least one of three properties: 1) a well-defined closed boundary; 2) a different appearance from their surroundings; 3) \KGtwo{uniqueness within the image}.
Building on the terminology from~\cite{Alexe}, we refer to our task as \emph{pixel objectness}. We use this name to distinguish our task from the related problems of salient object detection (which seeks only the most attention-grabbing foreground object) and region proposals (which seeks a ranked list of candidate object-like regions).

\KGtwo{It is important to acknowledge that some images prompt differences in human perception about what constitutes the primary foreground object~\cite{ambiguous-truth}.  Furthermore, in current video segmentation benchmarks (cf.~Sec.~\ref{sec:results}) there are (implicitly) different assumptions made about whether an object must move to be considered foreground.  While the proposed work does not aim to resolve human perception ambiguities, we find in practice that our model can be successfully trained once on off-the-shelf data and generalize well to a series of other image and video datasets.}

\KG{The proposed approach consists of a two-stream CNN architecture that infers pixel objectness from appearance and motion.  
Below we first present the appearance stream (Sec.~\ref{sec:app_pixelobject}), then the motion stream (Sec.~\ref{sec:motion}), followed by a fusion layer that brings the two together (Sec.~\ref{sec:joint}).  
Pixel objectness is applicable to either images and video.  For images, we have only appearance to analyze, and the motion stream is bypassed. }

\subsection{Appearance Stream}\label{sec:app_pixelobject}


Given an RGB image or video frame$~\mathcal{I}$ of size $m \times n \times c$ as input, we formulate the task of generic object segmentation as densely labeling each pixel as either ``object" or ``background".  Thus the output of pixel objectness is a binary map of size $m \times n$.

For an individual image, the main idea is to train the system to predict pixel objectness using a mix of \emph{explicit} boundary-level annotations and \emph{implicit} image-level object category annotations.  From the former, the system will obtain direct information about image cues indicative of generic foreground object boundaries.   From the latter, it will learn object-like features across a wide spectrum of object types---but \emph{without} being told where those objects' boundaries are.

To this end, for the appearance stream we train a fully convolutional deep neural network for the foreground-background object labeling task.  We initialize the network using a powerful generic image representation learned from millions of images labeled by their object category, but lacking any foreground annotations.  Then, we fine-tune the network to produce dense binary segmentation maps, using relatively few images with pixel-level annotations originating from a small number of object categories.   

Since the pretrained network is trained to recognize thousands of objects, we hypothesize that its image representation has a strong notion of objectness built inside it, even though it never observes \emph{any} segmentation annotations. 
Meanwhile, by subsequently training with explicit dense foreground labels, we can steer the method to fine-grained cues about boundaries that the standard object classification networks have no need to capture.   This way, even if our model is trained with a limited number of object categories having pixel-level annotations, we expect it to learn generic representations helpful to pixel objectness.

Specifically, we adopt a deep network structure~\cite{chen14semantic} originally designed for multi-class semantic segmentation (see Sec.~\ref{sec:results} for more implementation details).  We initialize it with weights pre-trained on ImageNet, which provides a representation equipped to perform image-level classification for some 1,000 object categories.  Next, we take a modestly sized semantic segmentation dataset, and transform its dense semantic masks into binary object vs.~background masks, by fusing together all its 20 categories into a single supercategory (``generic object").  We then train the deep network (initialized for ImageNet object classification) to perform well on the dense foreground pixel labeling task. The loss is the sum of cross-entropy terms over each pixel in the output layer. Our model supports end-to-end training.

 \begin{figure}[t]
\centering
\renewcommand{\tabcolsep}{0pt}
  \captionsetup{ font={footnotesize}, skip=2pt}

\includegraphics[width=1\columnwidth]{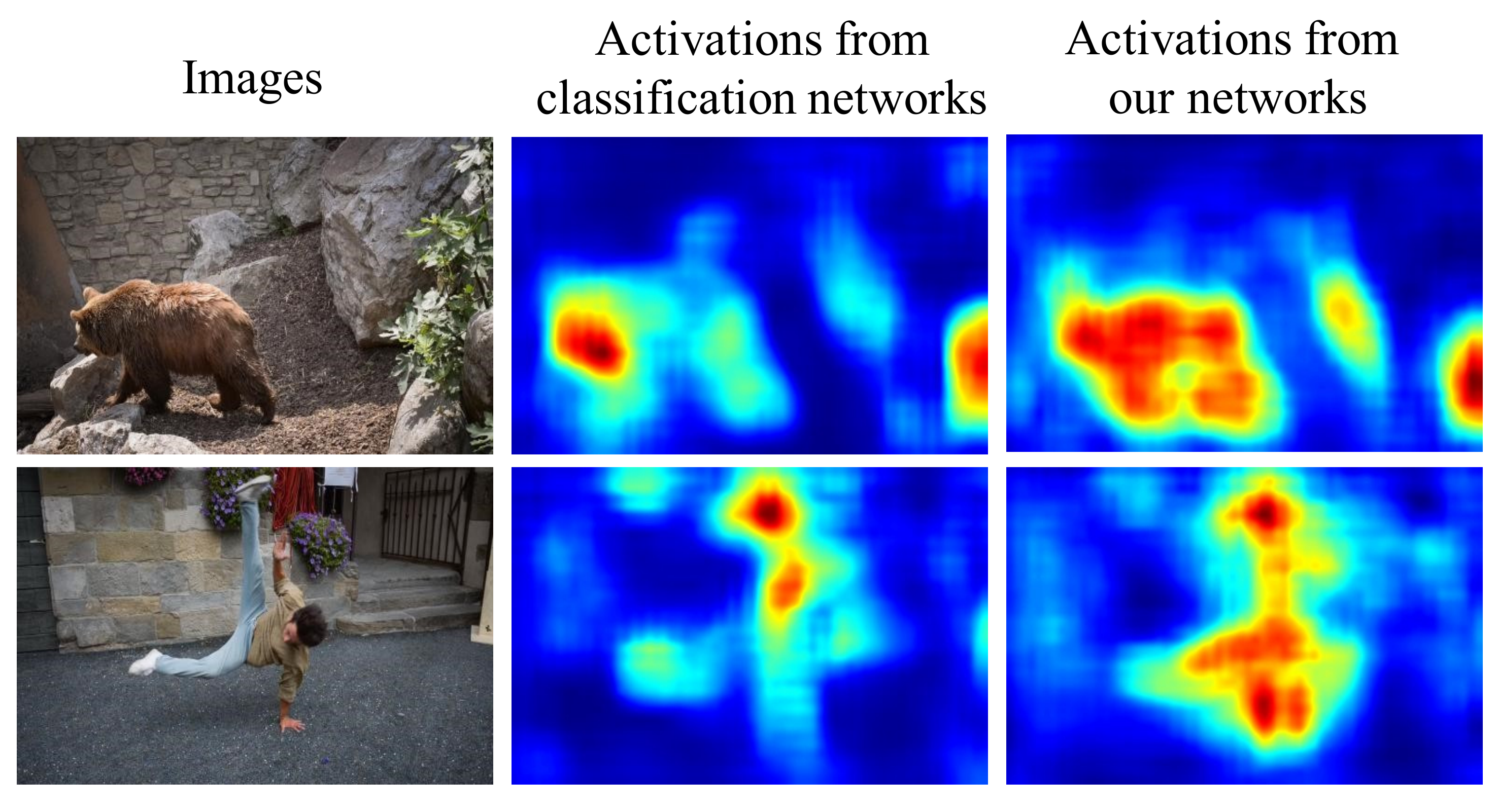}
\caption{Activation maps from a network (VGG~\cite{simonyan2014very}) trained for the classification task and our network which is fine-tuned with explicit dense foreground labels. We see that the classification network has already learned image representations that have some notion of objectness, but with poor ``over''-localization. Our network deepens the notion of objectness to pixels and captures fine-grained cues about boundaries.  Best viewed on pdf.}
\label{fig:activation}
\vspace{8pt}
\end{figure}

To illustrate this synergy, Fig.~\ref{fig:activation} shows activation maps from a network trained for ImageNet 
classification (middle) and from our network (right), by summing up feature 
responses from each filter in the last convolutional layer (pool5) for each spatial location. 
Although networks trained on a classification task never observe any segmentations, they can show high activation responses when object parts are present and low activation responses to 
stuff-like regions such as rocks and roads. Since the classification networks are trained with 
thousands of object categories, their activation responses are rather general. However, they are responsive to only fragments of the objects.   

After training with explicit dense foreground labels, our network is able to extend high activation responses from discriminative object parts to the entire object.  For example, in Fig.~\ref{fig:activation},  the classification 
network only has a high activation response on the bear's head, whereas  our pixel objectness network has a high response on the entire bear body; similarly for the person.  This supports our hypothesis that networks trained for classification tasks \BX{contain} a reasonable but incomplete basis for objectness, despite lacking any spatial annotations.  By subsequently training with explicit dense foreground labels, we can steer towards fine-grained cues about boundaries that the standard object classification networks have no need to capture.

\begin{figure*}[t]
\centering
\renewcommand{\tabcolsep}{0pt}
  \captionsetup{ font={footnotesize}, skip=2pt}
\includegraphics[width=2\columnwidth]{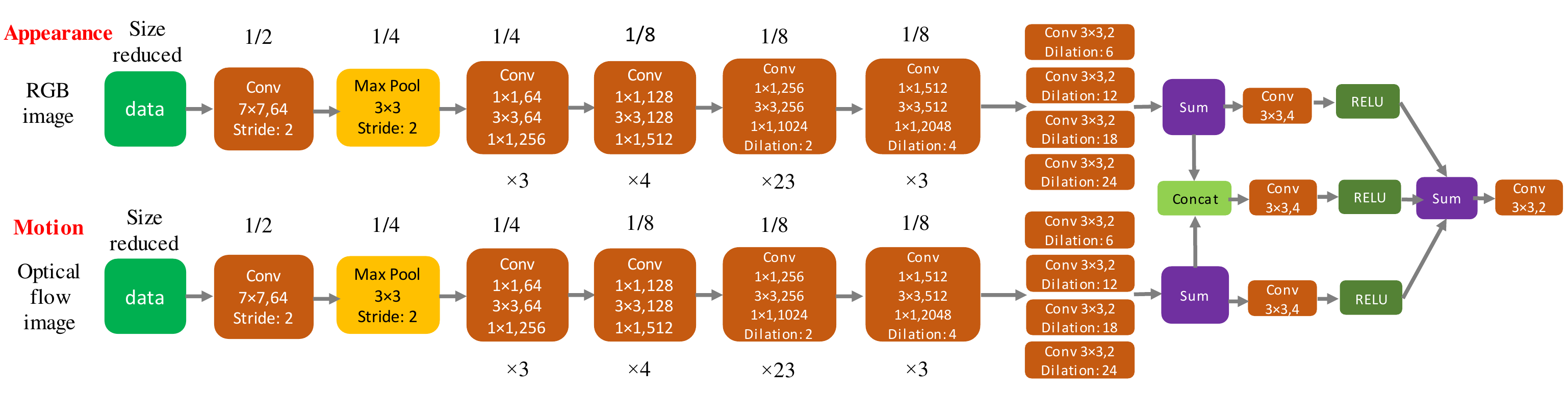}
\caption{Network structure for our segmentation model. Each convolutional layer except the first 7$\times$ 7  convolutional layer and our fusion blocks is a residual block~\cite{he2015deep}, adapted from ResNet-101. We show reduction in resolution at top of each box and the number of stacked convolutional layers in the bottom of each box.  \KG{To apply our model to images, only the appearance stream is used.}}
\label{fig:network}
\end{figure*}

\subsection{Motion Stream}
\label{sec:motion}

For the case of video segmentation, we have both the frame's appearance as well as its motion within the image sequence. 
Our complete video segmentation architecture consists of a two-stream network in which \KG{the appearance stream described thus far operates in parallel with a motion stream that processes the} optical flow image, then joins the two  in a fusion layer (see Fig.~\ref{fig:network}). We next discuss how to train a motion stream to densely predict pixel objectness from optical flow images only. \KG{Sec.~\ref{sec:joint} will explain how the two streams are merged.}  

The direct parallel to appearance-based pixel objectness discussed above would entail training the motion stream to map optical flow maps to video frame foreground maps.  However, an important practical catch to that solution is training data availability.  While ground truth foreground image segmentations are at least modestly available, datasets for video object segmentation masks are small-scale in deep learning terms, and primarily support evaluation.  For example, Segtrack-v2~\cite{rehg-iccv2013}, a commonly used benchmark dataset for video segmentation, contains only 14 videos with 1066 labeled frames. DAVIS~\cite{Perazzi2016} contains only 50 sequences with 3455 labeled frames. None contain enough labeled frames to train a deep neural network. 
Semantic video segmentation datasets like CamVid~\cite{BrostowFC:PRL2008} or Cityscapes~\cite{cityscapes} are somewhat larger, yet limited in object diversity due to a focus on street scenes and vehicles.   
A good training source for our task would have ample frames with human-drawn segmentations on a wide variety of foreground objects, and would show a good mix of static and moving objects.  No such large-scale dataset exists and creating one is non-trivial.

We propose a solution that leverages readily available \emph{image} segmentation annotations together with \emph{weakly annotated video} data to train our model.  In brief, we temporarily decouple the two streams of our model, and allow the appearance stream (Sec.~\ref{sec:app_pixelobject}) to hypothesize likely foreground regions in frames of a large video dataset annotated only by bounding boxes.  Since appearance alone need not produce perfect segmentations, we devise a series of filtering stages to generate high quality estimates of the true foreground.  These instances bootstrap pre-training of the optical flow stream, then the two streams are joined to learn the best combination from minimal human labeled training videos.

More specifically, given a video dataset with bounding boxes labeled for each object,\footnote{No matter the test videos, we rely on ImageNet Video data, which contains 3862 videos and 30 diverse objects.  See Sec.~\ref{sec:results}.} we ignore the category labels and map the boxes alone to each frame.  Then, we apply the appearance stream, thus far trained only from images labeled by their foreground masks, to compute a binary segmentation for each frame.

\begin{figure}[t]
\centering

\renewcommand{\tabcolsep}{0pt}
  \captionsetup{ font={footnotesize}, skip=2pt}
\includegraphics[width=1\columnwidth]{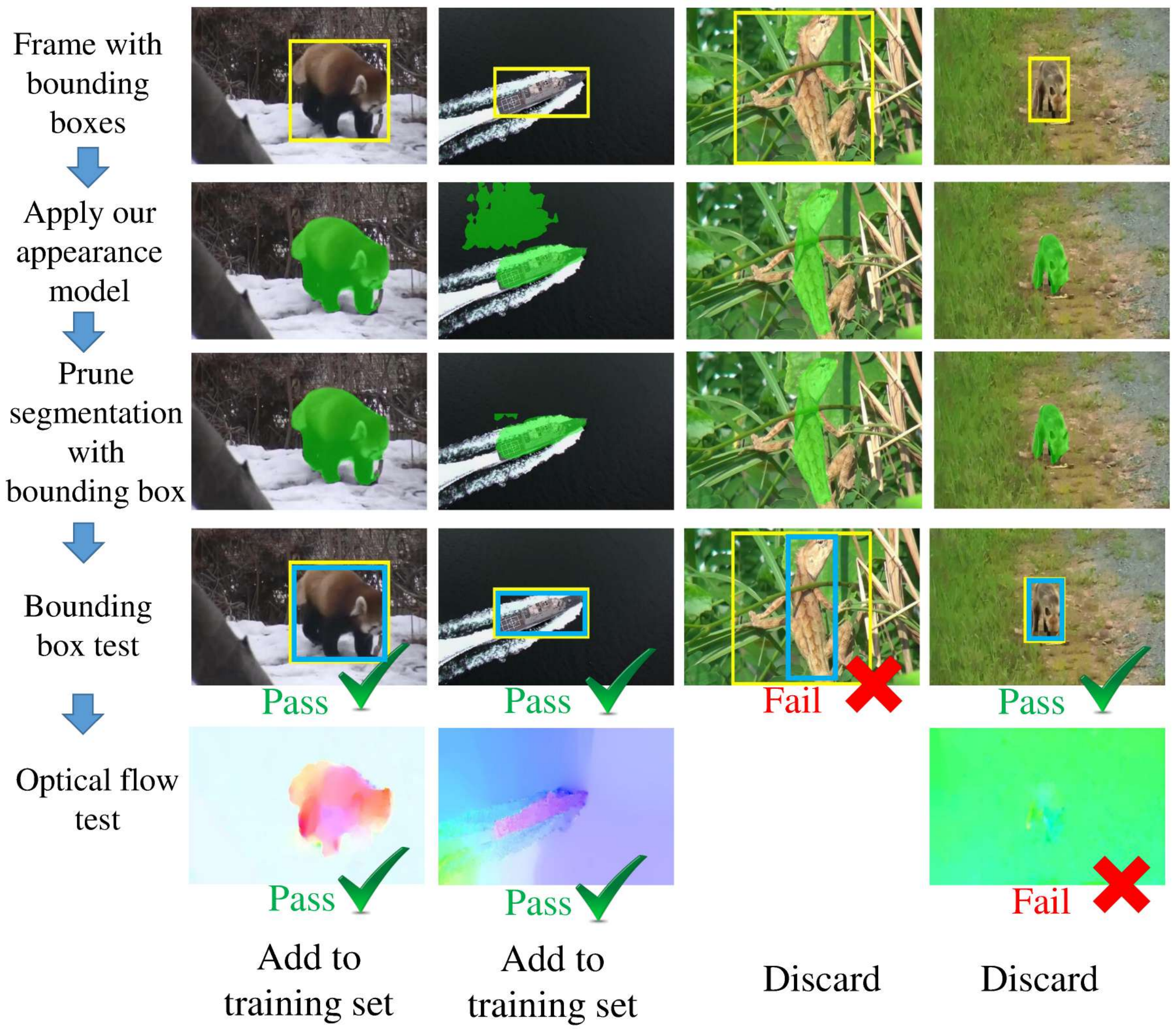}
\caption{Procedure to generate (pseudo)-ground truth segmentations. We first apply the appearance model to obtain initial segmentations (second row, with object segment in green) and then prune  by setting pixels outside bounding boxes as background (third row). Then we apply the bounding box test (fourth row, yellow bounding box is ground truth and blue bounding box is the smallest bounding box enclosing the foreground segment) and optical flow test (fifth row) to determine whether we add the segmentation to the motion stream's training set or discard it.  Best viewed in color.}

\label{fig:step}
\vspace{8pt}

\end{figure}

Next we deconflict the box and segmentation in each training frame.    
First, we refine the binary segmentation by setting all the pixels outside the bounding box(es) as background. 
Second, for each bounding box, we check if the the smallest rectangle that encloses all the foreground pixels overlaps with the bounding box by at least 75\%. Otherwise we discard the segmentation. 
Third, we discard regions where the box contains more than 95\% pixels labeled as foreground, based on the prior that good segmentations are rarely a rectangle, and thus probably the true foreground spills out beyond the box.  
Finally, we eliminate segments where object and background lack distinct optical flow,
so our motion model can learn from the desired cues.  Specifically, we compute the frame's optical flow using~\cite{liu2009beyond} and convert it to an RGB flow image~\cite{baker2011database}.
\KGtwo{We compute} a) the average value \KGtwo{per color channel} within the bounding box and b) the average value in a box \KGtwo{with the same center but enlarged by a factor of two, ignoring any portion outside of the image boundaries}. \KG{If their difference} exceeds 30, the frame and filtered segmentation are added to the training set. See Fig.~\ref{fig:step} for visual illustration of these steps.

To recap, bootstrapping from the preliminary appearance model, followed by bounding box pruning, bounding box tests, and the optical flow test, we can generate accurate per-pixel foreground masks for thousands of diverse moving objects---for which no such datasets exist to date.  Note that by eliminating training samples with these filters, we aim to reduce label noise for training.  However, at test time our system will be evaluated on standard benchmarks for which each frame is manually annotated (see Sec.~\ref{sec:results}).

With this data, we now turn to training the motion stream.  
Analogous to our strong generic appearance model, we also want to train a strong generic pixel objectness motion model that can segment foreground objects purely based on motion. 
Our motion model takes only optical flow as the input and is trained with automatically generated pixel level ground truth segmentations. In particular, we convert the raw optical flow to a 3-channel (RGB) color-coded optical flow image~\cite{baker2011database}.  We use this color-coded optical flow image as the input to the motion network. We again initialize our network with pre-trained weights from ImageNet classification~\cite{ILSVRC15}.  Representing optical flow using RGB flow images allows us to leverage the strong pre-trained initializations as well as maintain symmetry in the appearance and motion arms of the network.

An alternative solution might forgo handing the system optical flow, and instead input two raw consecutive RGB frames.  However, doing so would likely demand more training instances in order to discover the necessary cues.  Another alternative would directly train the joint model that combines both motion and appearance, whereas we first ``pre-train" each stream to make it discover convolutional features that rely on appearance or motion alone, followed by a fusion layer (below).  Our design choices are rooted in avoiding bias in training our model.  Since the (pseudo) ground truth comes from the initial appearance network, training jointly from the onset is liable to bias the network to exploit appearance at the expense of motion.   By feeding the motion model with only optical flow, we ensure our motion stream learns to segment objects from motion.

\subsection{Fusion Model}\label{sec:joint}

The final processing in our pipeline joins the outputs of the appearance and motion streams, and aims to leverage a whole that is greater than the sum of its parts.  We now describe how to train the joint model using both streams.

An object segmentation prediction is reliable if 1) either appearance or motion model alone predicts the object segmentation with very strong confidence or 2) their combination together predicts the segmentation with high confidence. This motivates the structure of our joint model. 

\bx{We implement the idea by creating three independent parallel branches: 1) We apply a 3$\times$3 convolution layer followed by a ReLU to the output of the appearance model. 2) We apply a 3$\times$3 convolution layer followed by a ReLU to the output of the motion model. 3) We concatenate the outputs of the appearance and motion models, and apply a 3$\times$3 convolution layer followed by a ReLU. We sum up the outputs from the three branches and apply a 3$\times$3 convolution layer to obtain the final prediction. See Fig. \ref{fig:network}. }

As discussed above, we do not fuse the two streams in an early stage because we want them both to have strong independent predictions.  We can then train the fusion model with very limited annotated video data, without overfitting. \CR{In the absence of large volumes of video segmentation training data, precluding a complete end-to-end training, our strategy of decoupling the individual streams and training works very well in practice.}

\section{Results}\label{sec:results}

We first present pixel objectness results on image segmentation (Sec.~\ref{sec:results_pixel_objectness}) and two applications that benefit from predicting pixel objectness (Sec.~\ref{sec:results_pixel_app}). Then we show results on video segmentation (Sec.~\ref{sec:results_video}).

\subsection{Results on Image Segmentation}\label{sec:results_image}
We evaluate pixel objectness by comparing it to \KG{16} recent methods in the literature, and also examine its utility for two applications: \bx{image retrieval} and image retargeting.

\noindent {\bf Datasets:} We use three datasets \iccvf{which are commonly used to evaluate foreground object segmentation in images:}
\begin{itemize}[leftmargin=*]
\item {\bf MIT Object Discovery:} This dataset consists of Airplanes, Cars, and Horses~\cite{rubinstein-cvpr2013}. It is most commonly used to evaluate weakly supervised segmentation methods. The images were primarily collected using internet search and the dataset comes with per-pixel ground truth segmentation masks. 
\item {\bf ImageNet-Localization:} We conduct a large-scale evaluation of our approach using ImageNet~\cite{ILSVRC15} ($\sim$1M images with bounding boxes, 3,624 classes). The diversity of this dataset lets us test the generalization abilities of our method. 
\item {\bf ImageNet-Segmentation:} This dataset contains 4,276 images from 445 ImageNet classes with pixel-wise ground truth from~\cite{guillaumin2014imagenet}. 

\end{itemize}

\noindent {\bf Baselines:} We compare to these state-of-the-art methods:
\begin{itemize}[leftmargin=*]
\item {\bf Saliency Detection:} We compare to four salient object detection methods~\cite{zhang2013saliency,jiangsaliency,zhao2015saliency,DeepSaliency}, selected for their efficiency and state-of-the-art performance. All these  methods are designed to produce a complete segmentation of the prominent object (vs.~fixation maps; see Sec. 5 of~\cite{zhang2013saliency}) and output continuous saliency maps, which are then thresholded by per image mean to obtain the segmentation, \KGtwo{which gave the best results.}

\item {\bf Object Proposals:} We compare with state-of-the-art region proposal algorithms, multiscale combinatorial grouping (MCG)~\cite{APBMM2014} and DeepMask~\cite{deepmask}. These methods output a ranked list of generic object segmentation proposals. The top ranked proposal in each image is taken as the final foreground segmentation for evaluation. We also compare with SalObj~\cite{secret} which uses saliency to merge multiple object proposals from MCG into a single foreground.

\item {\bf Weakly supervised joint-segmentation methods:} These approaches rely on additional weak supervision in the form of prior knowledge that all images in a given collection share a common object category~\cite{rubinstein-cvpr2013,chen_cvpr14,joulin-cvpr2010,joulin-cvpr2012,kim-iccv2011,TangCVPR14,suyog-cvpr2016}. Note that our method lacks this additional supervision.
\end{itemize}

\noindent {\bf Evaluation metrics:} Depending on the dataset, we use: \bx{1)} {\bf Jaccard Score:} Standard intersection-over-union (IoU) metric (\KGtwo{$\times$ 100)} between predicted and ground truth segmentation masks and \bx{2)} {\bf BBox-CorLoc Score:} Percentage of objects correctly localized with a bounding box according to PASCAL criterion (i.e IoU $>$ 0.5) used in \cite{TangCVPR14,ferrari_ijcv12}.

For MIT and ImageNet-Segmentation, we use the segmentation masks and evaluate using the Jaccard score. 
For ImageNet-Localization we evaluate with the BBox-CorLoc metric, following the setup from~\cite{TangCVPR14,ferrari_ijcv12},
\cc{which entails putting a tight bounding box around our method's output.}

{\bf Training details:} To generate the explicit boundary-level training data, we rely on the 1,464 PASCAL 2012 segmentation training images~\cite{Everingham2010} and the additional annotations of~\cite{BharathICCV2011}, for 10,582 total training images.  The 20 object labels are discarded and mapped instead to the single generic ``object-like'' (foreground) label for training. We train our model using the Caffe implementation of~\cite{chen14semantic}. We optimize with stochastic gradient with a mini-batch size of 10 images. A simple data augmentation through mirroring the input images is also employed. A base learning rate of 0.001 with a $1/10$th slow-down every 2000 iterations is used. We train the network for a total of 10,000 iterations; total training time was about 8 hours on a modern GPU.  \BO{We adopt the VGG~\cite{simonyan2014very} network structure for experiments on image segmentation in order to make fair  comparison with DeepSaliency~\cite{DeepSaliency}, which also adopts the VGG~\cite{simonyan2014very} network structure.}

\subsubsection{Foreground Object Segmentation Results}\label{sec:results_pixel_objectness}

\noindent {\bf MIT Object Discovery:} First we present results on the MIT dataset~\cite{rubinstein-cvpr2013}.   We do separate evaluation on the complete dataset and also a subset defined in~\cite{rubinstein-cvpr2013}. We compare our method with 13 existing state-of-the-art methods including saliency detection~\cite{zhang2013saliency,jiangsaliency,zhao2015saliency,DeepSaliency}, object proposal generation~\cite{APBMM2014,deepmask} plus merging~\cite{secret} and joint-segmentation~\cite{rubinstein-cvpr2013,chen_cvpr14,joulin-cvpr2010,joulin-cvpr2012,kim-iccv2011,suyog-cvpr2016}. We compare with author-reported results for the joint-segmentation baselines, and use software provided by the authors for the saliency and object proposal baselines.

\iccvf{Table~\ref{tab:results_mit} shows the results. Our proposed method outperforms several state-of-the-art saliency and object proposal methods---including recent deep learning techniques~\cite{zhao2015saliency,DeepSaliency,deepmask}---in three out of six cases, and is competitive with the best performing method in the others.}

Our gains over the joint segmentation methods are arguably even more impressive because our model simply segments a single image at a time---no weak supervision!---and still substantially outperforms all weakly supervised techniques. We stress that in addition to the weak supervision in the form of segmenting common object, the previous best performing method~\cite{suyog-cvpr2016} also makes use of a pre-trained deep network; we use strictly less total supervision than~\cite{suyog-cvpr2016} yet still perform better.    
Furthermore, most joint segmentation methods involve expensive steps such as dense correspondences~\cite{rubinstein-cvpr2013} or region matching~\cite{suyog-cvpr2016} which can take up to hours even for a modest collection of 100 images. In contrast, our method directly outputs the final segmentation in a single forward pass over the network and takes only 0.6 seconds per image for complete processing. \\ 

\begin{table}[t]
\centering
     	\captionsetup{width=0.48\textwidth, font={footnotesize}, skip=2pt}
     	   \scriptsize
     	   \setlength\tabcolsep{4pt}
   		\begin{tabular}{|c|c|c|c|c|c|c|}
   			\hline
     		\multirow{2}{*}{{\bf Methods}} & \multicolumn{3}{c|}{{\bf MIT dataset (subset)}} & \multicolumn{3}{c|}{{\bf MIT dataset (full)}} \\
\cline{2-7}
   			 & {\bf Airplane} & {\bf Car} & {\bf Horse} & {\bf Airplane} & {\bf Car} & {\bf Horse} \\
   			 \hline
   			 \hline
   							{\bf \# Images} & 82 & 89  & 93 & 470 & 1208 & 810 \\
   			\hline					
   \hline
	     \multicolumn{7}{|c|}{{\bf Joint Segmentation}} \\
   \hline
                            Joulin et al. \cite{joulin-cvpr2010} & 15.36 & 37.15  & 30.16 & n/a & n/a & n/a \\
\hline
   							Joulin et al. \cite{joulin-cvpr2012} & 11.72 & 35.15  & 29.53 & n/a & n/a & n/a \\
\hline
   							Kim et al. \cite{kim-iccv2011} & 7.9 &  0.04 & 6.43  & n/a & n/a & n/a \\
\hline
   							Rubinstein et al. \cite{rubinstein-cvpr2013} & 55.81 & 64.42 & 51.65  & 55.62 & 63.35  & 53.88 \\
\hline
   							Chen et al. \cite{chen_cvpr14} & 54.62 & 69.2  & 44.46   & 60.87 & 62.74 &  60.23 \\
\hline
   						Jain et al.~\cite{suyog-cvpr2016} & 58.65 & 66.47  & 53.57   & 62.27 &  65.3 & 55.41 \\
   \hline
   \hline
   	   
   	    \multicolumn{7}{|c|}{{\bf Saliency}} \\
   \hline
							    Jiang et al. \cite{jiangsaliency} & 37.22 & 55.22  & 47.02 & 41.52 & 54.34 & 49.67 \\
   \hline
      							Zhang et al. \cite{zhang2013saliency} & 51.84 & 46.61  & 39.52 & 54.09 & 47.38 & 44.12 \\
     \hline
						      DeepMC ~\cite{zhao2015saliency} & 41.75  & 59.16   & 39.34  & 42.84 &	58.13 &	41.85 \\  
     \hline
						      DeepSaliency~\cite{DeepSaliency} & 69.11 & 83.48  & 57.61 & {\bf 69.11} &	83.48	& {\bf 67.26} \\ 
     \hline
     \hline
        	    \multicolumn{7}{|c|}{{\bf Object Proposals}} \\
        \hline
				 MCG \cite{APBMM2014} & 32.02 & 54.21  & 37.85 & 35.32 & 52.98 & 40.44 \\
		\hline 
			    DeepMask~\cite{deepmask} & {\bf 71.81} & 67.01  & 58.80 &    68.89 &	65.4 &	62.61 \\
		\hline
				 SalObj \cite{secret} & 53.91 & 58.03  & 47.42 &  55.31 &	55.83	& 49.13 \\
\hline
\hline
  {\bf Ours} & 66.43 & {\bf 85.07}  & {\bf 60.85} & 66.18 & {\bf 84.80} & 64.90 \\
\hline
   				\end{tabular}
   				\caption{Quantitative results on MIT Object Discovery dataset. Our method outperforms several state-of-the-art methods for saliency detection, object proposals, and joint segmentation. (Metric: Jaccard score.)}
  		\label{tab:results_mit}
\end{table}

\noindent {\bf ImageNet-Localization:} Next we present results on the ImageNet-Localization dataset. This involves testing our method on about 1 million images from 3,624 object categories. This also lets us test how generalizable our method is to unseen categories, i.e., those for which the method  sees no \KGtwo{segmented} foreground examples during training.  \KGtwo{While our use of standard ImageNet pretraining means that our system has been exposed to the \emph{categories} (image-level labels), for nearly all of them, it has never seen them \emph{segmented}.}

Table~\ref{tab:results_imagenet} (left) shows the results. When doing the evaluation over all categories, we compare our method with five methods which report results on this dataset~\cite{Alexe,TangCVPR14,suyog-cvpr2016} or are scalable enough to be run at this large scale~\cite{jiangsaliency,APBMM2014}. We see that our method significantly improves the state-of-the-art.  The saliency and proposal methods~\cite{jiangsaliency,Alexe,APBMM2014} result in much poorer segmentations. Our method also significantly outperforms the joint segmentation approaches~\cite{TangCVPR14,suyog-cvpr2016}, which are the current best performing methods on this dataset. In terms of the actual number of images, our gains translate into correctly segmenting 42,900 more images than~\cite{suyog-cvpr2016} (which, like us, leverages ImageNet features) and 83,800 more images than~\cite{TangCVPR14}. This reflects the overall magnitude of our gains over state-of-the-art baselines.

\begin{table}[t]
	\centering
	\scriptsize
	\setlength{\tabcolsep}{0.3em}
	\captionsetup{width=0.48\textwidth, font={footnotesize}, skip=2pt}
	\begin{tabular}{|c|c|c|}
		\hline
		\multicolumn{3}{|c|}{{\bf ImageNet-Localization dataset}} \\
		\hline
		\hline
		
		 & All & Non-Pascal \\
		\hline
		\hline
		\# Classes & 3,624 & 3,149 \\
		\hline
		\# Images & 939,516 & 810,219 \\
		\hline
		\hline
		
		Alexe et al. \cite{Alexe} & 37.42 & n/a \\
		\hline
		Tang et al. \cite{TangCVPR14}  &  53.20 & n/a \\
		\hline
		Jain et al.~\cite{suyog-cvpr2016}& 57.64 & n/a \\
		\hline
		\hline
		Jiang et al. \cite{jiangsaliency} & 41.28 & 39.35 \\
		\hline
		MCG \cite{APBMM2014} & 42.23 & 41.15 \\
		\hline
				\hline
				
		Ours & {\bf 62.12} & {\bf 60.18} \\
		\hline
	\end{tabular}
	\qquad
		\begin{tabular}{| c | c |}
			\hline
			\multicolumn{2}{|c|}{{\bf ImageNet-Segmentation dataset}} \\
			\hline		
			\hline
			
			Jiang et al. \cite{jiangsaliency} & 43.16  \\
			\hline	
			Zhang et al. \cite{zhang2013saliency} & 45.07  \\
			\hline	
			DeepMC ~\cite{zhao2015saliency} & 40.23  \\
			\hline	
			DeepSaliency~\cite{DeepSaliency} &   62.12 \\
			\hline	
			\hline
			MCG \cite{APBMM2014} & 39.97  \\
			\hline		
			DeepMask~\cite{deepmask} & 58.69   \\
			\hline
			SalObj~\cite{secret} & 41.35   \\
			\hline 
			Guillaumin et al.~\cite{guillaumin2014imagenet} & 57.3 \\
			\hline
			\hline
			Ours & {\bf 64.22} \\
			\hline			
		\end{tabular}
	
			\caption{Quantitative results on ImageNet localization and segmentation datasets. Results on ImageNet-Localization (left) show that the proposed model outperforms several state-of-the-art methods and also generalizes very well to unseen object categories (Metric: BBox-CorLoc). It also outperforms all methods on the ImageNet-Segmentation dataset (right) showing that it produces high-quality object boundaries (Metric: Jaccard score).}
	\label{tab:results_imagenet}
\vspace{5pt}
\end{table}

\begin{figure*}[t]
	\centering
	\renewcommand{\tabcolsep}{0pt}
	\captionsetup{width=1\textwidth, font={footnotesize}, skip=2pt}
	\begin{tabular}{c}
		ImageNet Examples from Pascal Categories \\
		\includegraphics[keepaspectratio=true,scale=0.37]{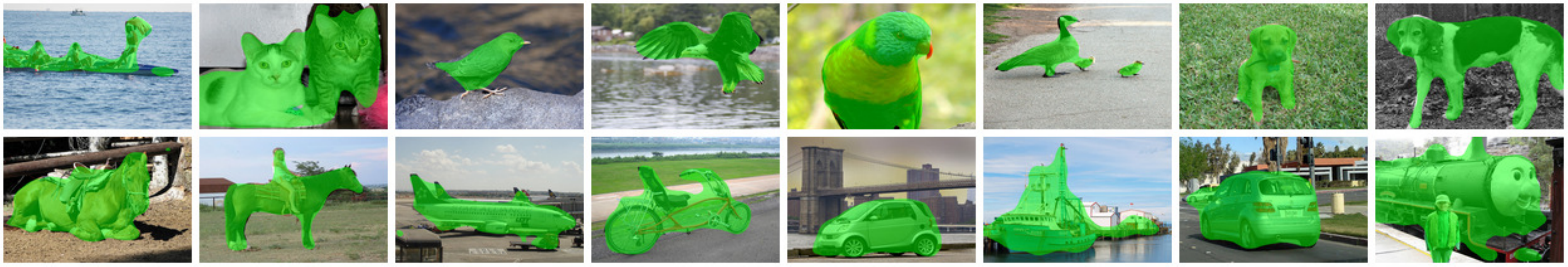} \\
		\hline
		ImageNet Examples from Non-Pascal Categories (unseen) \\
		\includegraphics[keepaspectratio=true,scale=0.37]{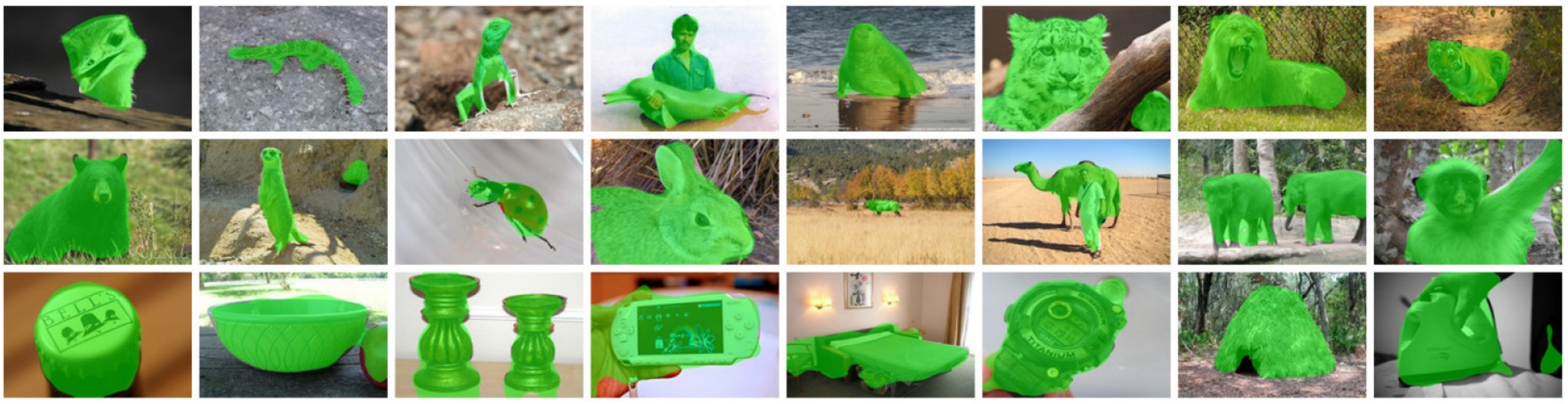} \\
		\hline
		Failure cases \\
		\includegraphics[keepaspectratio=true,scale=0.37]{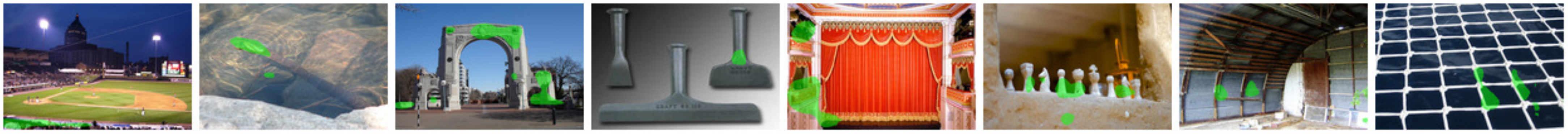} \\
	\end{tabular}
	\caption{Qualitative results: We show qualitative results on images belonging to PASCAL (top) and Non-PASCAL (middle) categories. Our segmentation model generalizes remarkably well even to those categories which were unseen in any foreground mask during training (middle rows). Typical failure cases (bottom) involve scene-centric images where it is not easy to clearly identify foreground objects (best viewed on pdf). }
	\label{fig:qual_res_pixel}
\end{figure*}

Does our learned segmentation model  only recognize foreground objects that it has seen during training, or can it generalize to unseen object categories? Intuitively, ImageNet has such a large number of diverse categories that this gain would not have been possible if our method was only over-fitting to the 20 seen PASCAL categories.  To empirically verify this intuition, we next exclude those ImageNet categories which are directly related to the PASCAL objects, by matching the two datasets' synsets.  This results in a total of 3,149 categories which are exclusive to ImageNet (``Non-PASCAL'').   See Table~\ref{tab:results_imagenet} (left) for the data statistics.

We see only a very marginal drop in performance; our method still significantly outperforms both the saliency and object proposal baselines. This is an important result, because during training the segmentation model \emph{never saw any dense object masks for images in these categories}.  Bootstrapping from the pretrained weights of the VGG-classification network, our model is able to learn a transformation between its prior belief on what looks like an object to complete dense foreground segmentations. \\

\noindent {\bf ImageNet-Segmentation: }
Finally, we measure the pixel-wise segmentation quality on a large scale.  For this we use the ground truth masks provided by~\cite{guillaumin2014imagenet} for 4,276 images from 445 ImageNet categories. \iccvf{The current best reported results are from the segmentation propagation approach of~\cite{guillaumin2014imagenet}. We found that DeepSaliency~\cite{DeepSaliency} and DeepMask~\cite{deepmask} further improve it. Note that like us, DeepSaliency~\cite{DeepSaliency} also trains with PASCAL~\cite{Everingham2010}. DeepMask~\cite{deepmask} is trained with a much larger COCO~\cite{LinECCV14coco} dataset. Our method outperforms all methods, significantly improving the state-of-the-art (see Table~\ref{tab:results_imagenet} (right)). This shows that our method not only generalizes to thousands of object categories but also produces high quality object segmentations.}

\begin{figure}[t!]
        \centering
        \captionsetup{width=0.48\textwidth, font={footnotesize}, skip=2pt}
        \includegraphics[width=\columnwidth]{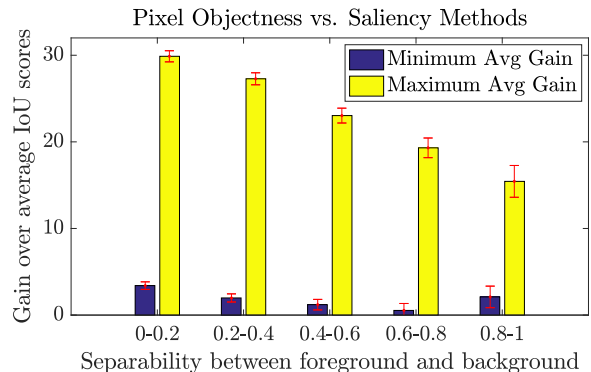}
        \caption{Pixel objectness vs. saliency methods: performance gains grouped using foreground-background separability scores. \bx{Error bars indicate standard error.}
On the x-axis, lower values mean that the objects are less salient and thus difficult to separate from background. On the y-axis, \bx{we plot the maximum and minimum average gains (\KGtwo{averaged over all test images}) of pixel objectness with \KGtwo{four} saliency methods.} }
        \label{fig:saliency_th}
\vspace{8pt}
\end{figure}

\noindent {\bf Pixel objectness vs.~saliency:} Salient object segmentation methods can potentially fail in cases where the foreground object does not stand out from the background. On the other hand, pixel objectness is designed to find objects even if they are not salient. To verify this hypothesis, we ranked all the images in the ImageNet-Segmentation dataset~\cite{guillaumin2014imagenet} by the appearance overlap between the foreground object and background. For this, we make use of the ground-truth segmentation to compute a 30-bin RGB color histogram for foreground and background respectively. We then compute cosine distance between the normalized histograms to measure how similar their distributions are and use that as a measure of \emph{separability} between foreground and background.

\begin{figure}[t!]
        \centering
        \renewcommand{\tabcolsep}{0pt}
        \captionsetup{width=0.48\textwidth, font={footnotesize}, skip=2pt}
        \begin{tabular}{c}
                Images \\
                \includegraphics[width=0.95\columnwidth]{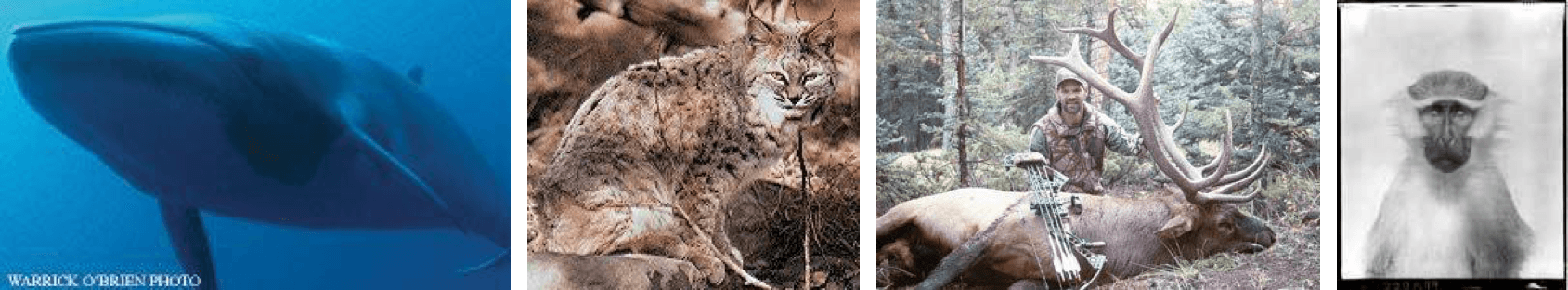} \\
                DeepSaliency~\cite{DeepSaliency} \\
                \includegraphics[width=0.95\columnwidth]{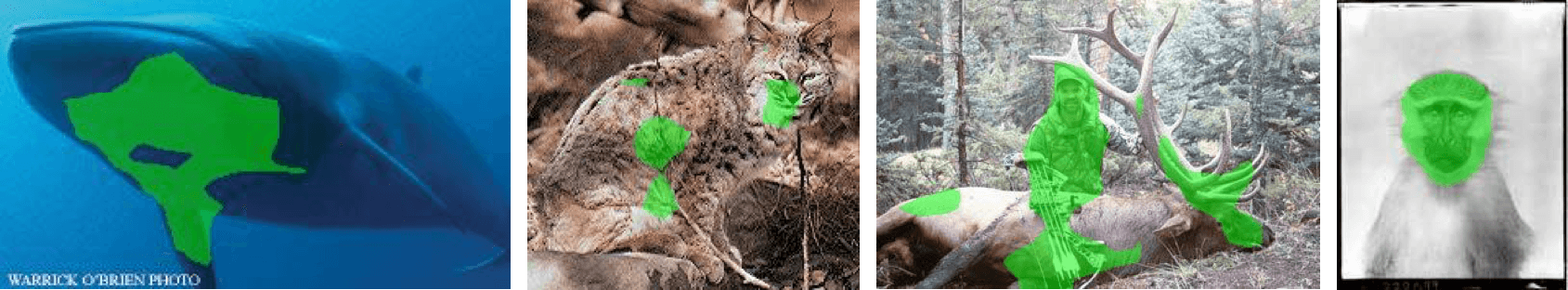} \\
                Ours \\
                \includegraphics[width=0.95\columnwidth]{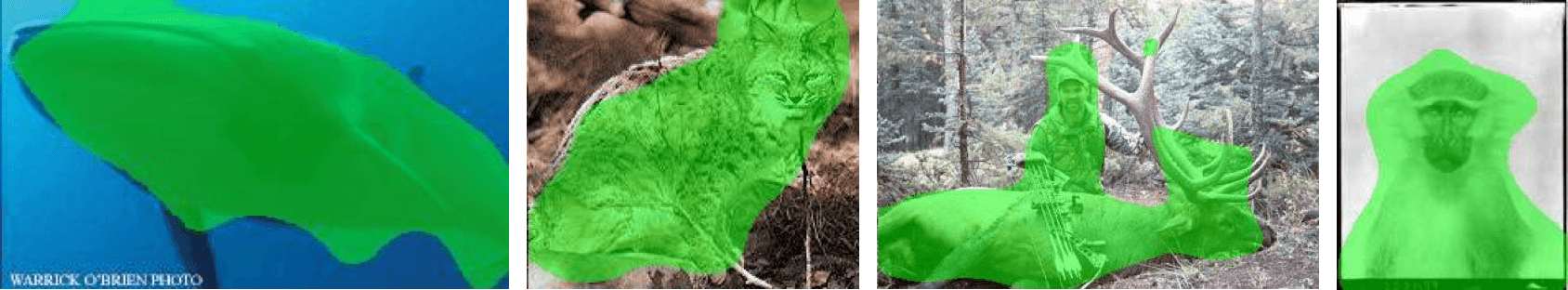} \\

        \end{tabular}
        \caption{Visual comparison for our method and the best performing saliency method, DeepSaliency~\cite{DeepSaliency}, which can fail when an object does not ``stand out" from background.  Best viewed on pdf. }
        \label{fig:qual_sal}
\vspace{8pt}
\end{figure}

Figure~\ref{fig:saliency_th} groups different images based on their separability scores and shows the minimum and maximum \KGtwo{average} gains of our method over four state-of-the-art saliency methods~\cite{zhang2013saliency,jiangsaliency,zhao2015saliency,DeepSaliency} for each group. \KGtwo{In other words, first we average the IoU scores over all test images in a given separability range, then we determine the \bbx{minimum and maximum} gain versus the averages of all four baselines.}
Lower separability score means that the foreground and background strongly overlaps and hence objects are not salient. First, we see that our method has positive gains across all groups showing that it outperforms all other saliency methods in every case. Secondly, we see that our gains are higher for lower separability scores. This demonstrates that the saliency methods are much weaker when foreground and background are not easily separable. On the other hand, pixel objectness works well irrespective of whether the foreground objects are salient or not. Our average gain over DeepSaliency~\cite{DeepSaliency} is 3.4\% IoU score on the subset obtained by thresholding at 0.2 (1320 images) as opposed to 2.1\% IoU score over the entire dataset.

Fig.~\ref{fig:qual_sal} visually illustrates this. Even the best performing saliency method~\cite{DeepSaliency} fails in cases where an object does not stand out from the background. In contrast, pixel objectness successfully finds complete objects in these images.  

\vspace{5pt}
\noindent {\bf Qualitative results:} \bx{Fig.~\ref{fig:qual_res_pixel}} shows qualitative results for ImageNet from both PASCAL and Non-PASCAL categories. Pixel objectness accurately segments foreground objects from both sets. The examples from the Non-PASCAL categories highlight its strong generalization capabilities. \iccvf{We are able to segment objects across scales and appearance variations, including multiple objects in an image. It can segment even man-made objects, which are especially distinct from the objects in PASCAL (see supp. for more examples). The bottom row shows failure cases. Our model has more difficulty in segmenting scene-centric images where it is more difficult to clearly identify foreground objects.}

\subsubsection{Impact on Downstream Applications}\label{sec:results_pixel_app}

Next we report results leveraging pixel objectness for two downstream tasks on images.
Dense pixel objectness has many applications. \iccvf{Here we explore how it can assist in image retrieval and content-aware image retargeting, both of which demand a single, high-quality estimate of the foreground object region.} 
\vspace{5pt}

\noindent\textbf{Object-aware image retrieval: } First, we consider how pixel objectness foregrounds can assist in image retrieval.  A retrieval system  accepts a query image containing an object, and then the system returns a ranked list of images that contain the same object. This is a valuable application, for example, to allow object-based online product search. Typically retrieval systems extract image features from the entire query image. This can be problematic, however, because it might retrieve images with similar background, especially when the object of interest is small.  We aim to use pixel objectness to restrict the system's attention to the foreground object(s) as opposed to the entire image. 

To implement the idea, we first run pixel objectness. In order to reduce false positive segmentations, we keep the largest connected foreground region if it is larger than $6\%$ of the overall image area. Then we crop the smallest bounding box enclosing the foreground segmentation and extract features from the entire bounding box. If no foreground is found (which occurs in roughly 17\% of all images), we extract image features from the entire image. The method is applied to both the query and database images. To rank database images, we explore two image representations. The first one uses only the image features extracted from the bounding box, and the second concatenates the features from the original image with those from the bounding box.

To test the retrieval task, we use the ILSVRC2012~\cite{ILSVRC15} validation set, which contains 50K images and $1,000$ object classes, with $50$ images per class. As an evaluation metric, we use mean average precision (mAP). We extract VGGNet~\cite{simonyan2014very} features and use cosine distance to rank retrieved images. We compare with two baselines 1) \textbf{Full image}, which ranks images based on features extracted from the entire image, and 2) \textbf{Top proposal} (TP), which ranks images based on features extracted from the top ranked MCG~\cite{APBMM2014} proposal.  For our method and the Top proposal baseline, we examine two image representations. The first directly uses the features extracted from the region containing the foreground or the top proposal (denoted \textbf{FG}). The second representation concatenates the extracted features with the image features extracted from the entire image (denoted \textbf{FF}).

Table~\ref{tab:result} shows the results. Our method with FF yields the best results. Our method outperforms both baselines for many ImageNet classes. 
We observe that our method performs extremely well on object-centric classes such as animals, but has limited improvement upon the baseline on scene-centric classes (lakeshore, seashore etc.).  To verify our hypothesis, we isolate the results on the first 400 object classes of ImageNet, which contain mostly object-centric classes, as opposed to scene-centric objects. On those first 400 object classes, our method outperforms both baselines by a larger margin. 
This demonstrates the value of our method at retrieving objects, which often contain diverse background and so naturally benefit more from accurate pixel objectness. \BO{Please see supp. for more analysis.}

\begin{table}[t]

\captionsetup{width=0.48\textwidth, font={footnotesize}, skip=2pt}
           \scriptsize
           \setlength\tabcolsep{4pt}
\centering
{\scriptsize
\hspace*{-0.1in}
\begin{tabular}{ |c|c|c|c|c|c|}
 \hline

Method& Ours(FF) &Ours(FG)& Full Img  & TP (FF)~\cite{APBMM2014} & TP (FG)~\cite{APBMM2014}  \\ \hline
All & \bf{0.3342}&0.3173 &  0.3082&0.3102&0.2092 \\ \hline
\BX{Obj-centric}  &\bf{0.4166}&0.4106  &0.3695&0.3734 &0.2679 \\ \hline

\end{tabular}}
\caption{Object-based image retrieval performance on ImageNet. We report average precision on the entire validation set (top), and on the first 400 categories (bottom), which are \BX{mostly object-centric classes}. }

\label{tab:result}
\vspace{5pt}
\end{table}

\vspace{5pt}

\noindent\textbf{Foreground-aware image retargeting:} As a second application, we explore how pixel objectness can enhance image retargeting. The goal is to adjust the aspect ratio or size of an image without distorting its important visual concepts. We build on the popular Seam Carving algorithm~\cite{avidan2007seam}, which eliminates the optimal irregularly shaped path, called a seam, from the image via dynamic programming.  In~\cite{avidan2007seam}, the energy is defined in terms of the image gradient magnitude. However, the gradient is not always a sufficient energy function, especially when important visual content is non-textured or the background is textured.

Our idea is to protect semantically important visual content based on foreground segmentation.  
To this end, we consider a simple adaption of Seam Carving.  We define an energy function based on high-level semantics rather than low-level image features alone. Specifically, we first predict pixel objectness, and then we scale the gradient energy $g$ within the foreground segment(s) by $(g +1) \times 2$.

\begin{figure}[t]
  \centering
  \renewcommand{\tabcolsep}{0pt}
  \captionsetup{font={footnotesize}, skip=2pt}

    \includegraphics[width=0.98\columnwidth]{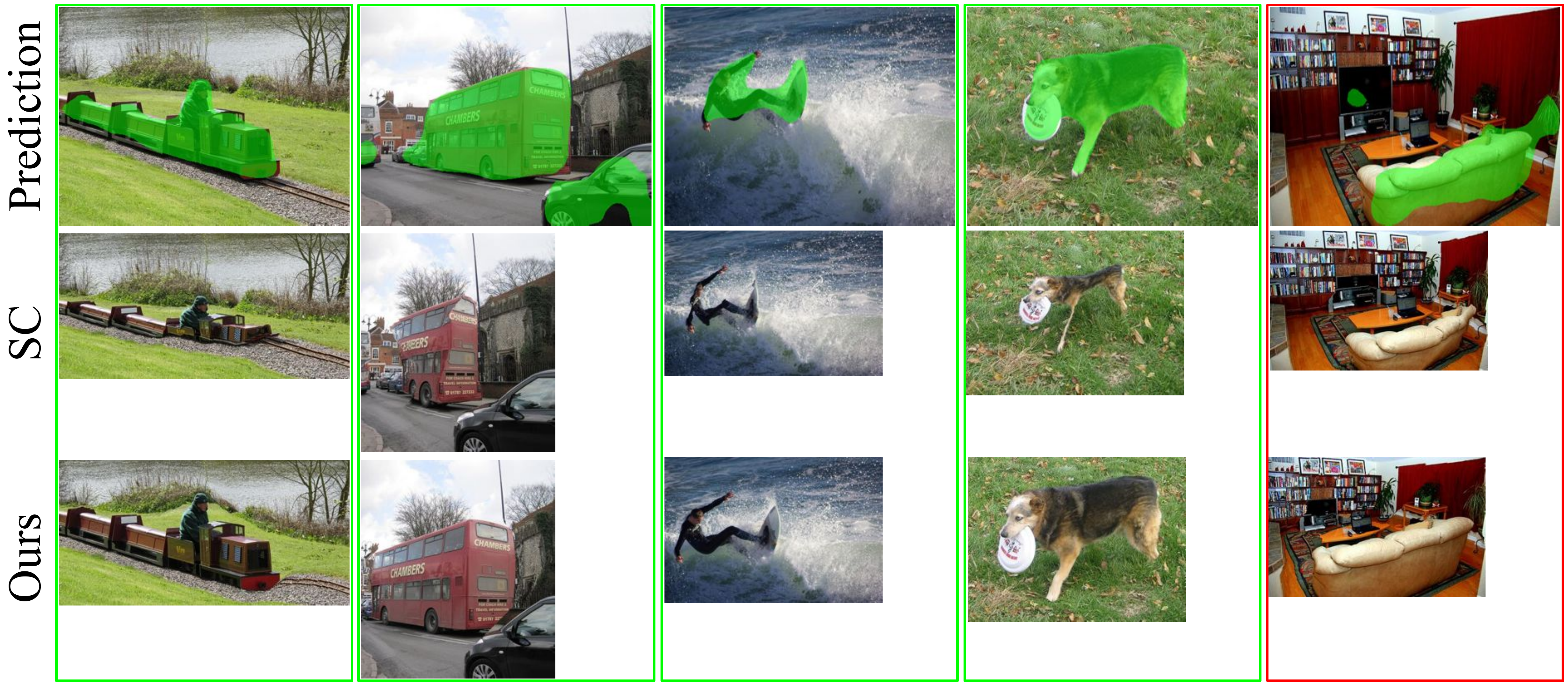} \\
   \caption{Leveraging pixel objectness for foreground aware image retargeting. See supp. for more examples.  Best viewed on pdf.}
\label{fig:seam_example}
\vspace{8pt}
\end{figure}

We use a random subset of 500 images from \KGtwo{MS COCO}\cite{LinECCV14coco} for experiments. Figure~\ref{fig:seam_example} shows example results. For reference, we also compare with the original Seam Carving (SC) algorithm \cite{avidan2007seam} that uses image gradients as  the energy function. 
Thanks to the proposed foreground segmentation, our method successfully preserves the important visual content (e.g., train, bus, human and dog) while reducing the content of the background. The baseline produces images with important objects distorted, because gradient strength is an inadequate indicator for perceived content, especially when background is textured. The rightmost column is a failure case for our method on a scene-centric image that does not contain any salient objects.

\KGtwo{In human subject studies on MTurk, we find that our results are preferred to the baseline's 39\% of the time, and the two methods tie 13\% of the time.  When the human subject is a vision researcher familiar with retargeting---but not involved in this project---those scores increase to 78\% and 9\%, respectively.  See supp.~for details.}

\subsection{Results on Video Segmentation}\label{sec:results_video}

Pixel objectness can predict high quality object segmentations and generalize very well to thousands of unseen object categories for image segmentation. We next show, when jointly trained with motion, our method also  improves the state-of-the-art results for \KG{automatically} segmenting generic objects in videos.

\vspace{5pt}
\noindent {\bf Datasets and metrics:} We evaluate on three challenging video object segmentation datasets: DAVIS~\cite{Perazzi2016}, YouTube-Objects~\cite{prest2012learning,suyog-eccv2014,tang-cvpr2013}, and SegTrack-v2~\cite{rehg-iccv2013}. To measure accuracy we again use the standard Jaccard score \KGtwo{($\times$ 100)}.  
The datasets are:

\begin{itemize}[leftmargin=*]
\item {\bf DAVIS~\cite{Perazzi2016}:} the latest and most challenging video object segmentation benchmark consisting of 50 high quality video sequences of diverse object categories with $3,455$  densely annotated, pixel-accurate frames. The videos are unconstrained in nature and contain challenges such as occlusions, motion blur, and appearance changes. Only the prominent moving objects are annotated in the ground-truth. 
\item {\bf YouTube-Objects~\cite{prest2012learning,suyog-eccv2014,tang-cvpr2013}:} consists of 126 challenging web videos from 10 object categories with more than 20,000 frames and is commonly used for evaluating video object segmentation. We use the subset defined in~\cite{tang-cvpr2013} and the ground truth provided by~\cite{suyog-eccv2014} for evaluation.
\item {\bf SegTrack-v2~\cite{rehg-iccv2013}:} one of the most common benchmarks for video object segmentation consisting of 14 videos with a total of $1,066$ frames with pixel-level annotations.  For videos with multiple objects with individual ground-truth segmentations, we treat them as a single foreground for evaluation.

\end{itemize}

\noindent \KGtwo{Note that these existing video datasets have slightly different (implicit) notions of what constitutes foreground objects, as reflected in their video selections and ground truth annotations.  In particular, objects in SegTrack and DAVIS exhibit motion, whereas in YouTube-Objects they need not exhibit motion.  Our approach is capable of handling either, and we do not tailor any design choices differently per dataset to allow it to do so.}

\setlength{\tabcolsep}{0.32em} 
{\renewcommand{\arraystretch}{1.2}
\begin{table*}[t]
	\centering
\scriptsize
	\begin{tabular}{|c|cccccccc|ccccc|ccc|}
		\hline
		\multicolumn{17}{|c|}{DAVIS: Densely Annotated Video Segmentation dataset (50 videos)}  \\
		\hline
		\hline
		Methods & Flow-T & Flow-S & PM~\cite{bideau2016s} & FST~\cite{ferrari-iccv2013} & KEY~\cite{keysegments} & NLC~\cite{nlc}  &MPN~\cite{tokmakov2016learning} &ARP~\cite{koh2017primary} & HVS~\cite{grundmann-cvpr2010} & FCP~\cite{Perazzi_2015_ICCV}  & BVS~\cite{marki2016bilateral} & VPN~\cite{jampani2017video}  & MSK~\cite{khoreva2017learning} & Ours-A & Ours-M & Ours-J \\
		
		\hline 
       \CR{Human?} & No & No & No &No & No & No & No & No & Yes & Yes  & Yes & Yes & Yes  & No & No & No \\
		\hline
		\hline		
		Avg. IoU (\%)  & 42.95 & 30.22 & 43.4 & 57.5 & 56.9 & 64.1 & 69.7 & {\bf{76.3}} & 59.6 & 63.1 & 66.5 &  75 & {\bf 80.3} & 64.69 & 60.18 & 72.82 \\
		\hline
	\end{tabular}
	\caption{\footnotesize{Video object segmentation results on DAVIS dataset. We show the average accuracy over all 50 videos. Our method outperforms \KGtwo{5 of the 6} \bbx{fully
automatic} state-of-the-art methods. 
\CR{The best performing methods grouped by whether they require human-in-the-loop or not during segmentation are highlighted in bold.}  Metric: Jaccard score, higher is better. } \BO{Please see supp. for per video results.}}
	\label{davis-results}
\end{table*}

\noindent {\bf Baselines:} We compare with several state-of-the-art methods for each dataset as reported in the literature. Here we group them together based on whether they can operate in a fully automatic fashion (automatic) or require a human in the loop (semi-supervised) to do the segmentation:

\begin{itemize}[leftmargin=*]
\item {\bf Automatic methods:} Automatic video segmentation methods do not require any human involvement to segment new videos. Depending on the dataset, we compare with the following state of the art methods: FST~\cite{ferrari-iccv2013}, KEY~\cite{keysegments}, NLC~\cite{nlc}, COSEG~\cite{Tsai_ECCV_2016}, MPN~\cite{tokmakov2016learning}, and \bx{ARP~\cite{koh2017primary}.} All use some form of unsupervised motion or objectness cues to identify foreground objects followed by post-processing to obtain space-time object segmentations.
	

\item {\bf Semi-supervised methods:} Semi-supervised methods bring a human in the loop. They have some knowledge about the object of interest which is exploited to obtain the segmentation (e.g., a manually annotated first frame). We compare with the following state-of-the-art methods: HVS~\cite{grundmann-cvpr2010}, HBT~\cite{godec11a}, FCP~\cite{Perazzi_2015_ICCV}, IVID~\cite{Nagaraja_2015_ICCV}, HOP~\cite{suyog-eccv2014}, and BVS~\cite{marki2016bilateral}. The methods require different amounts of human annotation to operate, e.g.  HOP, BVS, and FCP make use of manual complete object segmentation in the first frame to seed the method; HBT requests a bounding box around the object of interest in the first frame; HVS, IVID require a human to constantly guide the algorithm whenever it fails. We also compare with three semi-supervised video segmentation based on deep learning: VPN~\cite{jampani2017video}, MSK~\cite{khoreva2017learning} and OSVOS~\cite{caelles2017one}.
\end{itemize}

\noindent Our method requires human annotated data only during training. At test time it operates in a fully automatic fashion.  
Thus, given a new video, we require equal effort as the automatic methods, and less effort than the semi-supervised methods.

Apart from these comparisons, we also examine some natural baselines and variants of our method:

\begin{itemize}[leftmargin=*]

\item {\bf Flow-thresholding (Flow-T):} To examine the effectiveness of motion alone in segmenting objects, we  adaptively threshold the optical flow in each frame using the flow magnitude.  Specifically, we compute the mean and standard deviation from the L2 norm of flow magnitude and use ``mean+unit std." as the threshold.
\item {\bf Flow-saliency (Flow-S):} Optical flow magnitudes can have large variances, hence we also try a variant which normalizes the flow by applying a saliency detection method~\cite{jiang2013saliency} to the flow image itself. We use average thresholding to obtain the segmentation.
\bx{\item {\bf Probabilistic model for flow (PM)~\cite{bideau2016s}:} We compare with a prior \KGtwo{method} that uses a probabilistic model~\cite{bideau2016s} to segment objects relying on motion cues only.}
\item {\bf Appearance model (Ours-A):} To quantify the role of appearance in segmenting objects, we obtain segmentations using only the appearance stream of our model. 
\item {\bf Motion model (Ours-M):} To quantify the role of motion, we obtain segmentations using only our motion stream.
\item {\bf Joint model (Ours-J):} Our complete joint model that learns to combine both motion and appearance together to obtain the final object segmentation.

\end{itemize}

\begin{table*}[t]
	\centering
	\footnotesize
	\begin{tabular}{|c|ccccc|cccc|ccc|}
		\hline
		\multicolumn{13}{|c|}{YouTube-Objects dataset (126 videos)}  \\
		\hline		
		\hline		
		Methods & Flow-T & Flow-S & PM~\cite{bideau2016s} & FST~\cite{ferrari-iccv2013} & COSEG~\cite{Tsai_ECCV_2016} & HBT~\cite{godec11a} &  HOP~\cite{suyog-eccv2014} & IVID~\cite{Nagaraja_2015_ICCV} & OSVOS~\cite{caelles2017one} & Ours-A & Ours-M & Ours-J \\
		
		\hline			
		\CR{Human?}& No & No & No & No  & No  & Yes & Yes & Yes & Yes & No & No & No \\
		\hline		
		\hline		
		airplane (6) & 18.27 & 33.32 &25.83 & 70.9 & 69.3 & 73.6 & 86.27 & {\bf 89} & 88.2 & {\bf 83.38} & 59.38 & 83.09 \\
		bird (6) & 31.63 & 33.74 &26.27 & 70.6 & {\bf 76} & 56.1 & 81.04 & 81.6 & {\bf 85.7}   & 60.89 & 64.06 & 63.01 \\
		boat (15) & 4.35 & 22.59 & 12.54& 42.5 & 53.5 & 57.8 & 68.59 & 74.2 & {\bf 77.5}& 72.62   & 40.21 & {\bf 72.70} \\
		car (7) & 21.93 & 48.63 &37.90&  65.2 & 70.4 & 33.9 & 69.36 & 70.9 &{\bf 79.6} & 74.50 & 61.32   & {\bf 75.49} \\
		cat (16) & 19.9 & 32.33 &30.01&  52.1 & 66.8 & 30.5 & 58.89 & 67.7 & {\bf 70.8} & {\bf 67.99} & 49.16 &  67.75 \\
		cow (20) & 16.56 & 29.11 &35.31&  44.5 & 49 & 41.8 & 68.56 & {\bf 79.1}    & 77.8         & 69.63 & 39.38 & {\bf 70.30} \\
		dog (27) & 17.8 & 25.43 & 36.4&   65.3 & 47.5 & 36.8 & 61.78 & 70.3  & {\bf 81.3}       & {\bf 69.10} & 54.79 &  67.64 \\
		horse (14) & 12.23 & 24.17 &28.09&  53.5 & 55.7 & 44.3 & 53.96 & 67.8& {\bf 72.8}     & 62.79 & 39.96 & {\bf 65.05} \\
		mbike (10) & 12.99 & 17.06 & 24.08& 44.2 & 39.5 & 48.9 & 60.87 & 61.5& {\bf 73.5}    & 61.92 & 42.95 & {\bf 62.22} \\
		train (5) & 18.16 & 24.21 &23.62&  29.6 & 53.4 & 39.2 & 66.33 & {\bf 78.2} & 75.7     & {\bf 62.82} & 43.13 & 62.30 \\
		\hline
		Avg. IoU (\%) & 17.38 & 29.05 &28.01 & 53.84 & 58.11 & 46.29 & 67.56 & 74.03 & {\bf 78.3}    & 68.57 & 49.43 & {\bf 68.95} \\
		\hline
	\end{tabular}
	\caption{\footnotesize{Video object segmentation results on YouTube-Objects dataset. We show the average performance for each of the 10 categories from the dataset. The final row shows an average over all the videos. Our method outperforms \KGtwo{all other unsupervised methods}, and \KGtwo{half of those that require human annotation} during segmentation. The best performing methods grouped by whether they require human-in-the-loop or not during segmentation are highlighted in bold. Metric: Jaccard score, higher is better. }}
	\label{youtube-results}
\end{table*}

\begin{table*}[t!]
	\centering
	\footnotesize
	\begin{tabular}{|c|cccccc|ccc|ccc|}
		\hline
		\multicolumn{13}{|c|}{SegTrack-v2 dataset (14 videos)}  \\
		\hline		
		\hline		
		Methods & Flow-T & Flow-S  &  PM~\cite{bideau2016s} & FST~\cite{ferrari-iccv2013} & KEY~\cite{keysegments} & NLC~\cite{nlc} & HBT~\cite{godec11a} & HVS~\cite{grundmann-cvpr2010} & MSK~\cite{khoreva2017learning}& Ours-A & Ours-M & Ours-J \\
		
		\hline	
	    \CR{Human?} & No & No & No & No & No & No & Yes & Yes & Yes & No & No & No \\
		\hline	
		Avg. IoU (\%) & 37.77 & 27.04 & 33.5 & 53.5 & 57.3 & {\bf 80\textsuperscript{*} } & 41.3 & 50.8 & {\bf 67.4} & 56.88 & 53.04 & 64.44 \\
		\hline	
	\end{tabular}
	\caption{\footnotesize{Video object segmentation results on SegTrack-v2. We show the average accuracy over all 14 videos. Our method outperforms \KGtwo{most} state-of-the-art methods, including the ones which actually require human annotation during segmentation. The best performing methods grouped by whether they require human-in-the-loop or not during segmentation are highlighted in bold. $^{*}$For NLC results are averaged over 12 videos as reported in their paper~\cite{nlc}, \KG{whereas all other methods are tested on all 14 videos.} Metric: Jaccard score, higher is better.} \BO{Please see supp. for per video results.}} 
	\label{segtrack-results}
\end{table*}

\begin{figure*}[t]
  \centering
  \footnotesize
  \renewcommand{\tabcolsep}{0pt}
  \captionsetup{width=1\textwidth, font={footnotesize}, skip=2pt}
   \begin{tabular}{cc}
    \multicolumn{2}{c}{\includegraphics[keepaspectratio=true,scale=0.093]{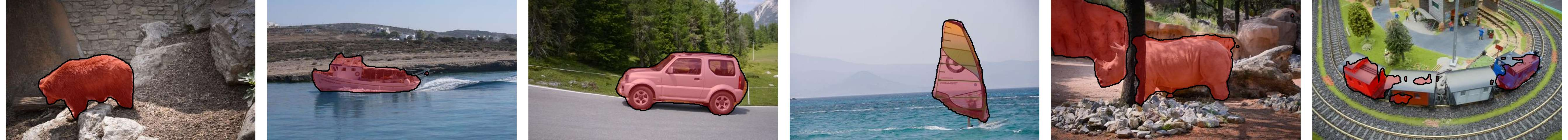}} \\
	\multicolumn{2}{c}{Appearance model (Ours-A)} \\    
    \multicolumn{2}{c}{\includegraphics[keepaspectratio=true,scale=0.093]{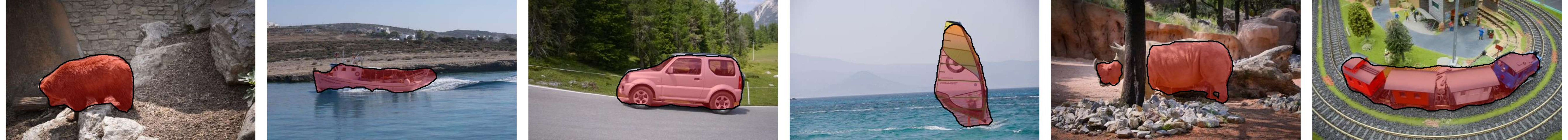}} \\    
	\multicolumn{2}{c}{Motion model (Ours-M)} \\    
    \multicolumn{2}{c}{\includegraphics[keepaspectratio=true,scale=0.093]{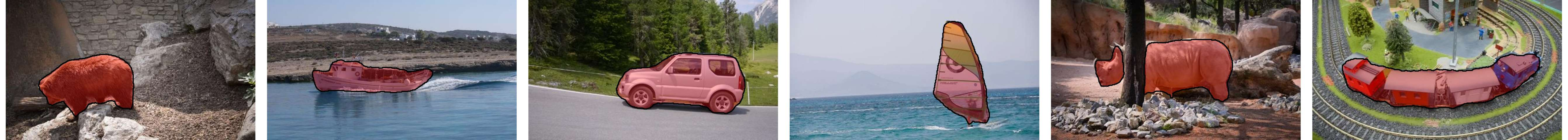}} \\
	\multicolumn{2}{c}{Joint model (Ours-J)} \\    
    \multicolumn{2}{c}{\includegraphics[keepaspectratio=true,scale=0.093]{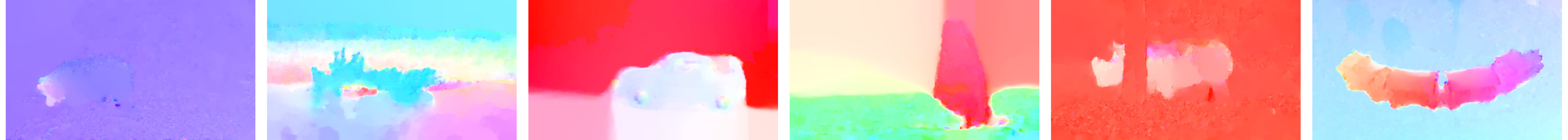}} \\    
	\multicolumn{2}{c}{Optical Flow Image} \\    
    \\
    {\small{\bf Ours vs. Automatic}} & {\small{\bf Ours vs. Semi-supervised}} \\
    \includegraphics[keepaspectratio=true,scale=0.093]{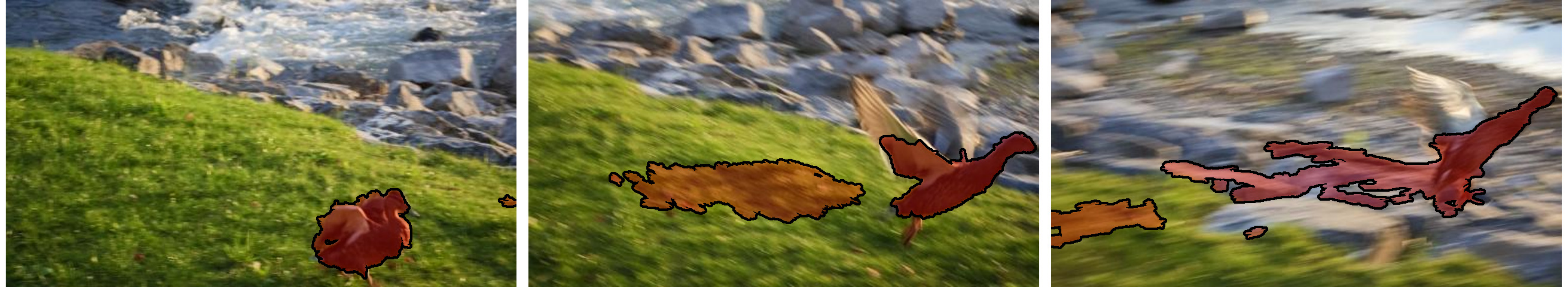} & \includegraphics[keepaspectratio=true,scale=0.093]{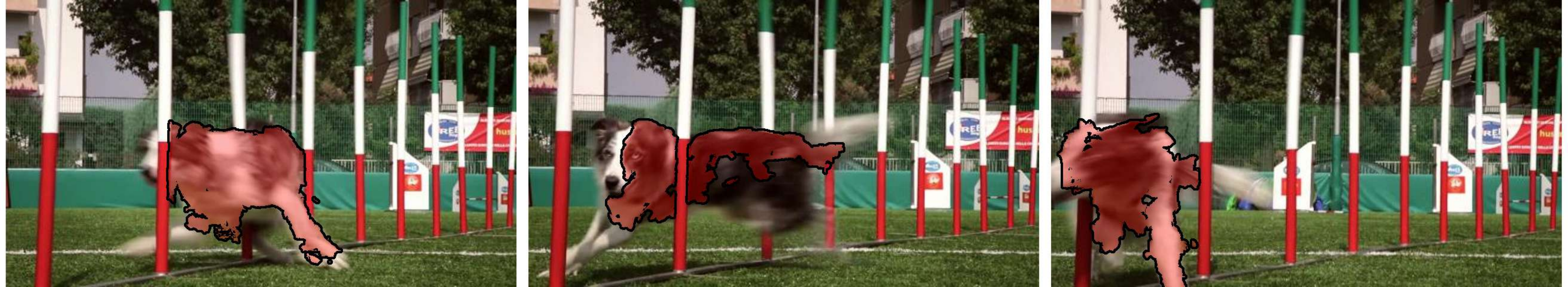} \\    
	 FST~\cite{ferrari-iccv2013} & BVS~\cite{marki2016bilateral} \\
    \includegraphics[keepaspectratio=true,scale=0.093]{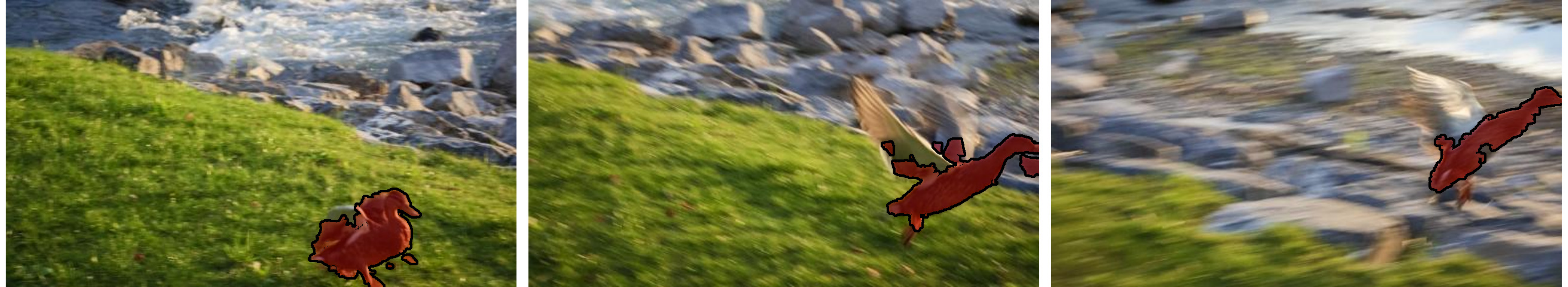} & \includegraphics[keepaspectratio=true,scale=0.093]{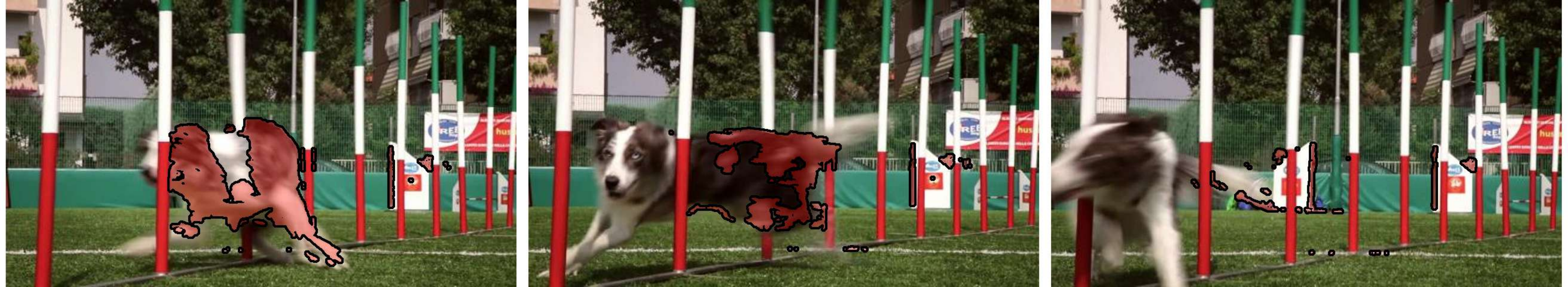} \\    
     NLC~\cite{nlc} & FCP~\cite{Perazzi_2015_ICCV} \\
    \includegraphics[keepaspectratio=true,scale=0.093]{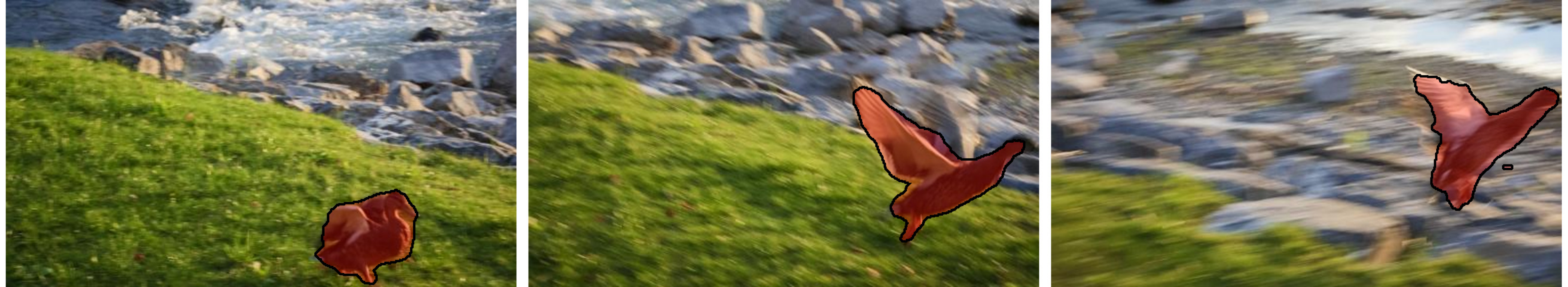} & \includegraphics[keepaspectratio=true,scale=0.093]{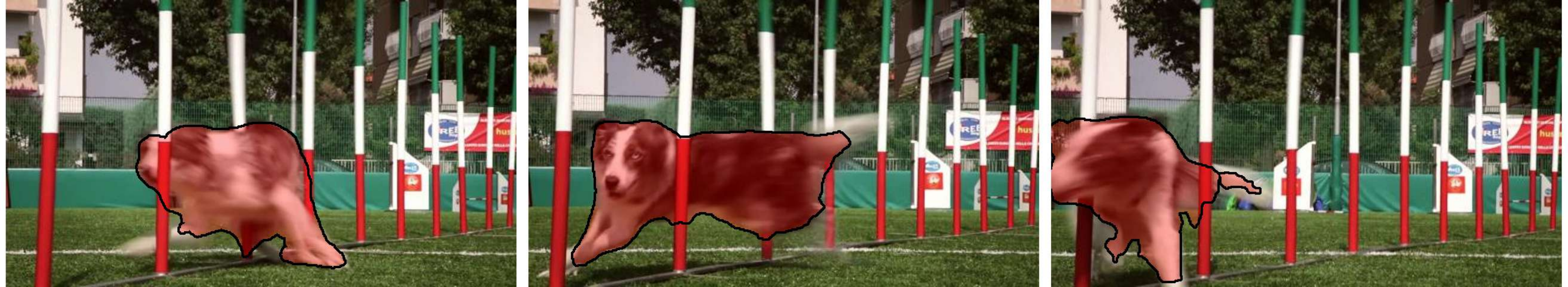} \\    
    \bx{Ours-J} & Ours-J \\
	\end{tabular}
	\caption{Qualitative results: The top half shows examples from our appearance, motion, and joint models along with the flow image which was used as an input to the motion network.  The bottom rows show visual comparisons of our method with automatic and semi-supervised baselines (best viewed on pdf and see text for the discussion). Videos of our segmentation results are available on the project website. }
	\label{fig:qual_res}
\end{figure*}

\noindent {\bf Implementation details:} 
As weak bounding box video annotations, we use the ImageNet-Video dataset~\cite{ILSVRC15}.  This dataset comes with a total of 3,862 training videos from 30 object categories with 866,870 labeled object bounding boxes from over a million frames. Post refinement using our ground truth generation procedure (see Sec.~\ref{sec:motion}), we are left with 84,929 frames with good pixel segmentations\footnote{\CR{Available for download on our project website.}} which are then used to train our motion model. \CR{We train each stream for a total of 20,000 iterations, use ``poly" learning rate policy (power = 0.9) with momentum (0.9) and weight decay (0.0005).  No post-processing is applied on the segmentations obtained from our networks. }
\bx{The DAVIS dataset is split into DAVIS-Train and DAVIS-Val. We train the joint model with DAVIS-Train, and report its performance on DAVIS-Val, SegTrack, and YouTube-Objects. We then train the joint model with DAVIS-Val, and report its performance on DAVIS-Train.}  \KGtwo{The latter flip ensures fairness when testing on DAVIS.}

\noindent {\bf Quality of training data:} To ascertain that the quality of training data we automatically generate for training our motion stream is good, we first compare it with a small amount of human annotated ground truth.  We randomly select 100 frames that passed both the bounding box and optical flow tests, and collect human-drawn segmentations on Amazon MTurk.  We first present crowd workers a frame with a bounding box labeled for each object, and then ask them to draw the detailed segmentation for all objects within the bounding boxes. Each frame is labeled by three crowd workers and the final segmentation is obtained by majority vote on each pixel. 
The results indicate that our strategy to gather pseudo-ground truth is effective.  On the 100 labeled frames, Jaccard overlap with the human-drawn ground truth is 77.8 (and 70.2 before pruning with bounding boxes).  


\noindent {\bf Quantitative evaluation:} We now present the quantitative comparisons of our method with several state-of-the-art methods and baselines, for each of the three datasets in turn.

\noindent {\bf DAVIS dataset:} Table~\ref{davis-results} shows the results, with \KG{baselines} that are the best performing methods taken from the benchmark results~\cite{Perazzi2016}. 
\bx{Our method achieves the second best performance among all the fully automatic methods.  The best  performing method ARP~\cite{koh2017primary}, proposed concurrently with our method, segments objects with an iterative augmentation and reduction process.}
Our method is significantly better than simple flow baselines. This supports our claim that even though motion contains a strong signal about foreground objects in videos, it is not straightforward to simply threshold optical flow and obtain those segmentations. A data-driven approach that learns to identify motion patterns indicative of objects as opposed to backgrounds or camera motion is required. 

The appearance and motion variants of our method themselves result in a very good performance. The performance of the motion variant is particularly \KGtwo{exciting}, knowing that it has no information about \KGtwo{the} object's appearance and purely relies on the flow signal. When combined together, the joint model results in a significant improvement, with an absolute gain of up to 11\% over \KGtwo{the} individual streams.

Our method is significantly better than \KGtwo{5 of the 6} fully automatic methods, which typically rely on motion alone to identify foreground objects. This illustrates the benefits of a unified combination of both motion and appearance. \KGtwo{Our} method also significantly outperforms several semi-supervised techniques, which require substantial human annotation on every video they process. The state-of-the-art human-in-the-loop algorithm MSK~\cite{khoreva2017learning} achieves better performance than ours. However, their method requires the first frame of the video to be manually segmented, \KG{whereas our method uses no human input.}

\noindent {\bf  YouTube-Objects dataset:} In Table~\ref{youtube-results} we see a similarly strong result on the YouTube-Objects dataset. Our method again outperforms the flow baselines and all the automatic methods by a significant margin. The publicly available code for NLC~\cite{nlc} runs successfully only on 9\% of the YouTube dataset (1725 frames); on those, its Jaccard score is 43.64\%. Our proposed model outperforms it by a significant margin of 25\%. Even among human-in-the-loop methods, we outperform all methods except IVID~\cite{Nagaraja_2015_ICCV} and OSVOS~\cite{caelles2017one}. However, both 
methods~\cite{Nagaraja_2015_ICCV,caelles2017one} require manual annotations. In particular, IVID~\cite{Nagaraja_2015_ICCV} requires a human to consistently track the segmentation performance and correct whatever mistakes the algorithm makes. This can take up to minutes of annotation time for each video. Our method uses zero human involvement but still performs competitively.

\noindent {\bf  Segtrack-v2 dataset:} In Table~\ref{segtrack-results}, our method outperforms all automatic methods except NLC~\cite{nlc} on Segtrack. While our approach significantly outperforms NLC~\cite{nlc} on the DAVIS dataset, NLC is exceptionally strong on this dataset.  Our relatively weaker performance could be due to the low quality and resolution of the Segtrack-v2 videos, making it hard for our network based model to process them. Nonetheless, our joint model still provides a significant boost over both our appearance and motion models, showing it again realizes the true synergy of motion and appearance. 

\noindent {\bf Qualitative evaluation:} Fig.~\ref{fig:qual_res} shows qualitative results. The top half shows visual comparisons between different components of our method including the appearance, motion, and joint models. We also show the optical flow image that was used as an input to the motion stream. These images help reveal the complexity of learned motion signals.  In the bear example, the flow is most salient only on the bear's head, still our motion stream alone is able to segment the bear completely. The boat, car, and sail examples show that even when the flow is noisy---including strong flow on the background---our motion model is able to learn about object shapes and successfully suppresses the background. The rhino and train examples show cases where the appearance model fails but when combined with the motion stream, the joint model produces accurate segmentations.  

The bottom half of Fig.~\ref{fig:qual_res} shows visual comparisons between our method and state-of-the-art automatic~\cite{ferrari-iccv2013,nlc} and semi-supervised~\cite{Perazzi_2015_ICCV,marki2016bilateral} methods. The automatic methods have a very weak notion about object's appearance; hence they completely miss parts of objects~\cite{nlc} or cannot disambiguate the objects from background~\cite{ferrari-iccv2013}. Semi-supervised methods~\cite{Perazzi_2015_ICCV,marki2016bilateral}, which rely heavily on the initial human-segmented frame to learn about object's appearance, start to fail as time elapses and the object's appearance changes considerably. In contrast, our method successfully learns to combine generic cues about object motion and appearance, segmenting much more accurately across all frames even in very challenging videos.

\section{Conclusions}

We proposed an end-to-end learning framework for segmenting generic objects in both images and videos.  Our results show that pixel objectness generalizes very well to thousands of unseen object categories for image segmentation. In addition, our method for video segmentation achieves \KGtwo{a true} synergy between motion and appearance and also addresses practical challenges in training a two-stream deep network. 
In the future, we plan to augment the framework with the ability to perform instance segmentation in both images and videos.

\vspace{10pt}
\noindent {\bf Examples, code, and pre-trained models available at:} \\ \url{http://vision.cs.utexas.edu/projects/pixelobjectness/} \\

\vspace{-20pt}
\section*{Acknowledgements}

\KGtwo{This research is supported in part by ONR YIP N00014-12-1-0754, an AWS Machine Learning Research Award, and the DARPA Lifelong Learning Machines project.  This material is based on research sponsored by the Air Force Research Laboratory and DARPA under agreement number FA8750-18-2-0126. The U.S. Government is authorized to reproduce and distribute reprints for Governmental purposes notwithstanding any copyright notation thereon. The authors thank the reviewers for their  suggestions.}

\vspace{-5pt}

\bibliographystyle{IEEEtran}
\bibliography{generic_object_extraction}


%

\vspace{-30pt}
\begin{IEEEbiography}
[{\includegraphics[width=1in,height=1.25in,clip,keepaspectratio]{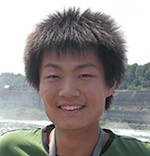}}]
{Bo Xiong}
received the BA degree in Computer Science and Mathematics from Connecticut College in 2013. He is currently a PhD student in the Department of Computer Science at the University of Texas at Austin. His research interests are in computer vision and deep learning.

\end{IEEEbiography}

\vspace{-45pt}
\begin{IEEEbiography}
[{\includegraphics[width=1in,height=1.25in,clip,keepaspectratio]{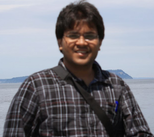}}]
{Suyog Dutt Jain}
received the PhD degree in computer science from the University of Texas at Austin in May 2017. He is currently a research scientist at CognitiveScale. His research interests include image and video segmentation, human computer interaction and crowdsourcing.

\end{IEEEbiography}
\vspace{-45pt}
\begin{IEEEbiography}
[{\includegraphics[width=1in,height=1.25in,clip,keepaspectratio]{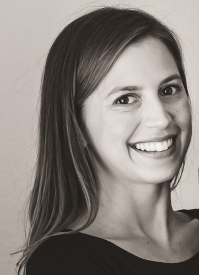}}]{Kristen Grauman}
is a Professor in the Computer Science Department at UT Austin.  She received her Ph.D. from MIT in 2007.  She is a recipient of the Sloan Research Fellowship, NSF CAREER and ONR Young Investigator awards, the 2013 PAMI Young Researcher Award, the 2013 IJCAI Computers and Thought Award, a 2013 Presidential Early Career Award for Scientists and Engineers (PECASE), and a 2017 Helmholtz Prize.  She and her collaborators were recognized with best paper awards at CVPR 2008, ICCV 2011, and ACCV 2016.  She currently serves as Associate Editor in Chief of PAMI and as a Program Chair of NIPS 2018.
\end{IEEEbiography}

\onecolumn

\section{Supplementary Materials}

In this section, we show:
 \begin{enumerate}
 
  \item more example image segmentations obtained by our pixel objectness model (referenced in \textbf{\emph{Section~\ref{sec:results_pixel_objectness}}} from the main paper)
  \item more object-aware image retrieval analysis (referenced in \textbf{\emph{Section~\ref{sec:results_pixel_app}}} from the main paper)
\bx{\item more details on human study for image retargeting (referenced in \textbf{\emph{Section~\ref{sec:results_pixel_app}}} from the main paper)}
  \item more foreground-aware image retargeting examples (referenced in \textbf{\emph{Fig.~\ref{fig:seam_example}}} from the main paper)
  \item Amazon Mechanical Turk interface used to collect human judgement for image retargeting (referenced in \textbf{\emph{Section~\ref{sec:results_pixel_app}}} from the main paper)
  \item Per video results for the DAVIS and Segtrack-v2 datasets (referenced in \textbf{\emph{Table~\ref{davis-results}}} and \textbf{\emph{Table~\ref{segtrack-results}}} from the main paper)
\end{enumerate}

\subsection{Pixel Objectness Qualitative Results } We show additional qualitative results from ImageNet dataset for our proposed pixel objectness model applied to images. Figures~\ref{fig:qual_res_seen1} and~\ref{fig:qual_res_seen2} show the qualitative results for ImageNet images which belong to PASCAL categories. Our method is able to accurately segment foreground objects, including cases with multiple foreground objects as well as the ones where the foreground objects are not highly salient.

Figures~\ref{fig:qual_res_unseen1},~\ref{fig:qual_res_unseen2}, and~\ref{fig:qual_res_unseen3} show qualitative results for those ImageNet images which belong to the non-PASCAL categories. Even though trained only on foregrounds from PASCAL categories, our method generalizes well. As can be seen, it can accurately segment foreground objects from completely disjoint categories, examples of which were never seen during training. Figure~\ref{fig:fail} shows more failure cases.

\begin{figure*}[h!]
  \centering
  \renewcommand{\tabcolsep}{0pt}
  \captionsetup{width=1\textwidth, font={footnotesize}, skip=2pt}
   \begin{tabular}{|c|}
      \hline
      {\bf Additional ImageNet Qualitative Examples from PASCAL Categories} \\
      \hline
    \includegraphics[width=0.9\columnwidth]{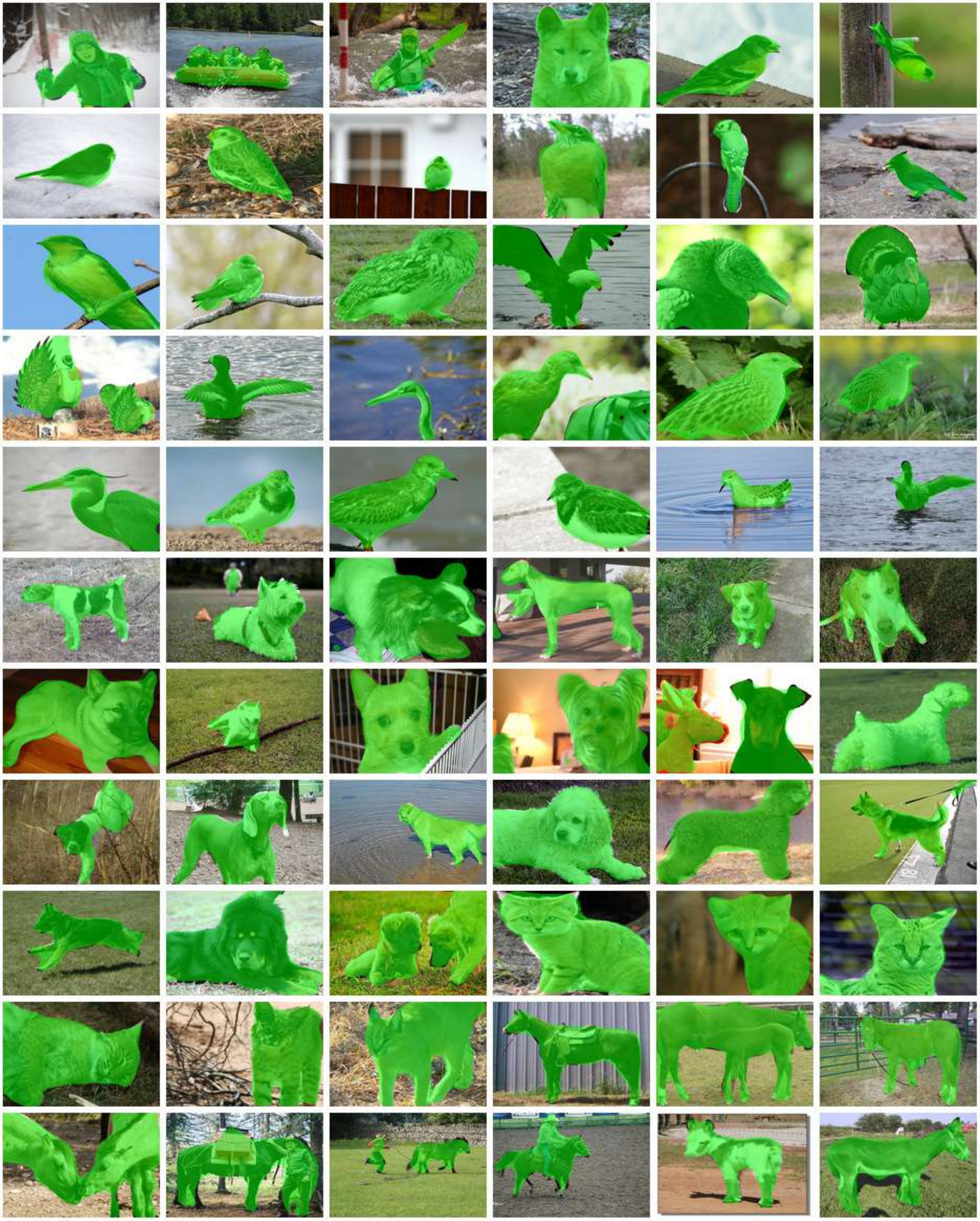} \\
	\hline
	\end{tabular}
		\caption{Qualitative results: We show example segmentations from ImageNet dataset obtained by our pixel objectness model. The segmentation results are shown with a green overlay. Our method is able to accurately segment foreground objects including cases where the objects are not highly salient. Best viewed in color.}
	  \label{fig:qual_res_seen1}
\end{figure*}

\begin{figure*}[h!]
  \centering
  \renewcommand{\tabcolsep}{0pt}
  \captionsetup{width=1\textwidth, font={footnotesize}, skip=2pt}
   \begin{tabular}{|c|}
      \hline
      {\bf Additional ImageNet Qualitative Examples from PASCAL Categories} \\
      \hline
    \includegraphics[width=0.9\columnwidth]{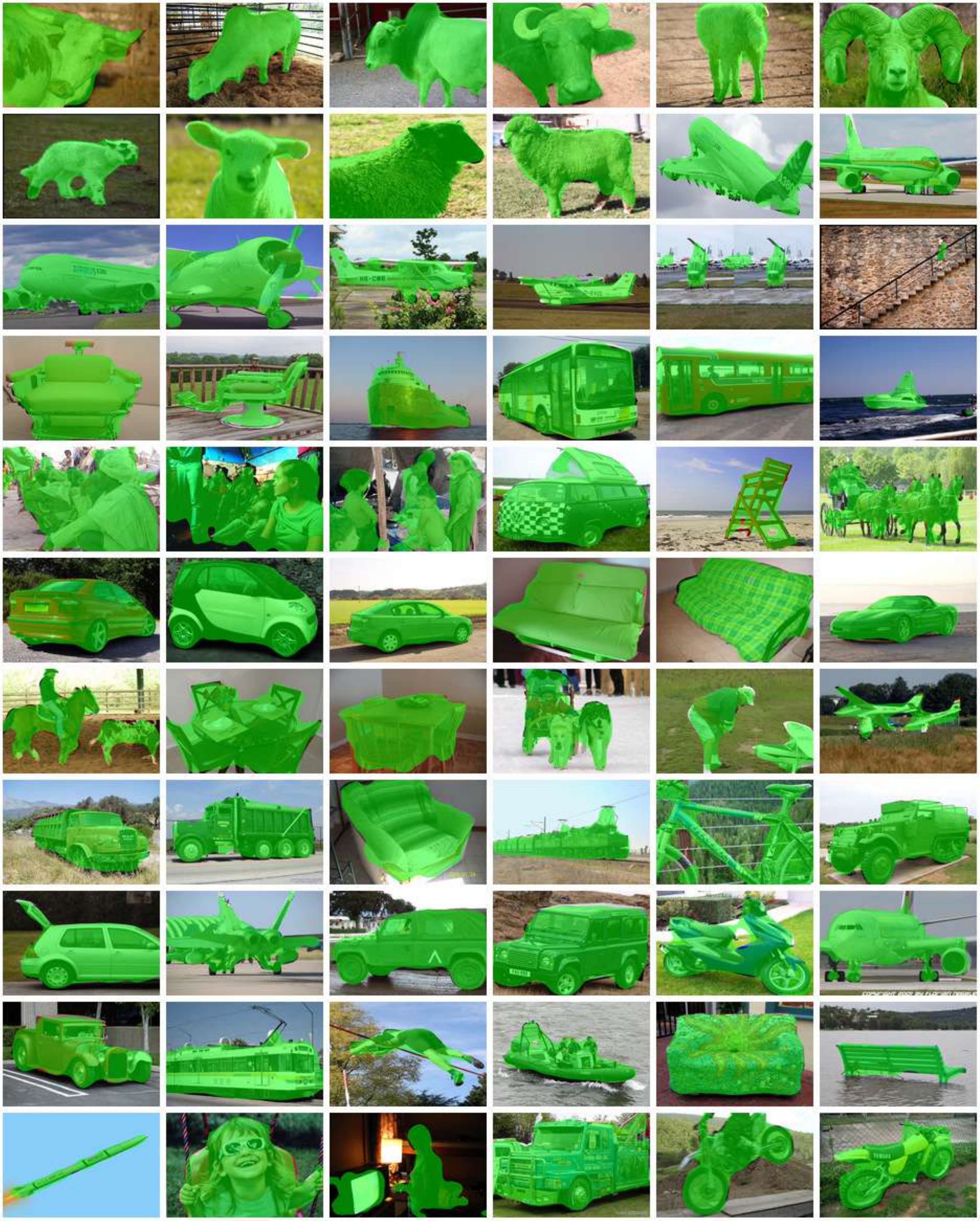} \\
	\hline
	\end{tabular}
	\caption{Qualitative results: We show example segmentations from ImageNet dataset obtained by our pixel objectness model on PASCAL Categories. The segmentation results are shown with a green overlay. Our method is able to accurately segment foreground objects including cases where the objects are not highly salient. Best viewed in color.}
	  \label{fig:qual_res_seen2}
\end{figure*}

\begin{figure*}[h!]
  \centering
  \renewcommand{\tabcolsep}{0pt}
  \captionsetup{width=1\textwidth, font={footnotesize}, skip=2pt}
   \begin{tabular}{|c|}
      \hline
      {\bf Additional ImageNet Qualitative Examples from Non-PASCAL (unseen) Categories} \\
      \hline
    \includegraphics[width=0.9\columnwidth]{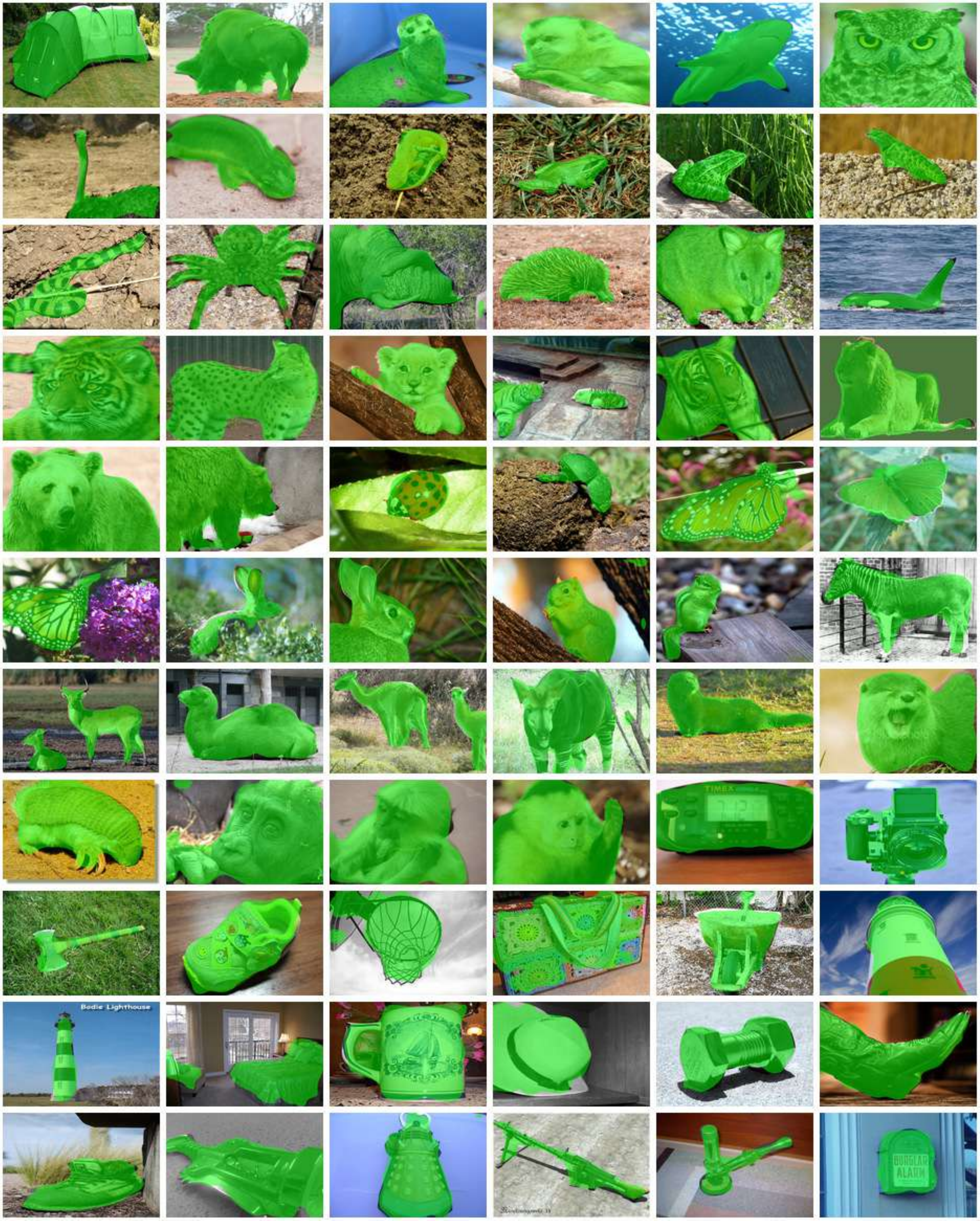} \\
	\hline
	\end{tabular}
	\caption{Qualitative results: We show example segmentations from ImageNet dataset obtained by our pixel objectness model on Non-PASCAL Categories. The segmentation results are shown with a green overlay. Our method generalizes remarkably well and is able to accurately segment foreground objects even for those categories which were never seen during training. Best viewed in color.}
	  \label{fig:qual_res_unseen1}
\end{figure*}

\begin{figure*}[h!]
  \centering
  \renewcommand{\tabcolsep}{0pt}
  \captionsetup{width=1\textwidth, font={footnotesize}, skip=2pt}
   \begin{tabular}{|c|}
      \hline
      {\bf Additional ImageNet Qualitative Examples from Non-PASCAL (unseen) Categories} \\
      \hline
    \includegraphics[width=0.9\columnwidth]{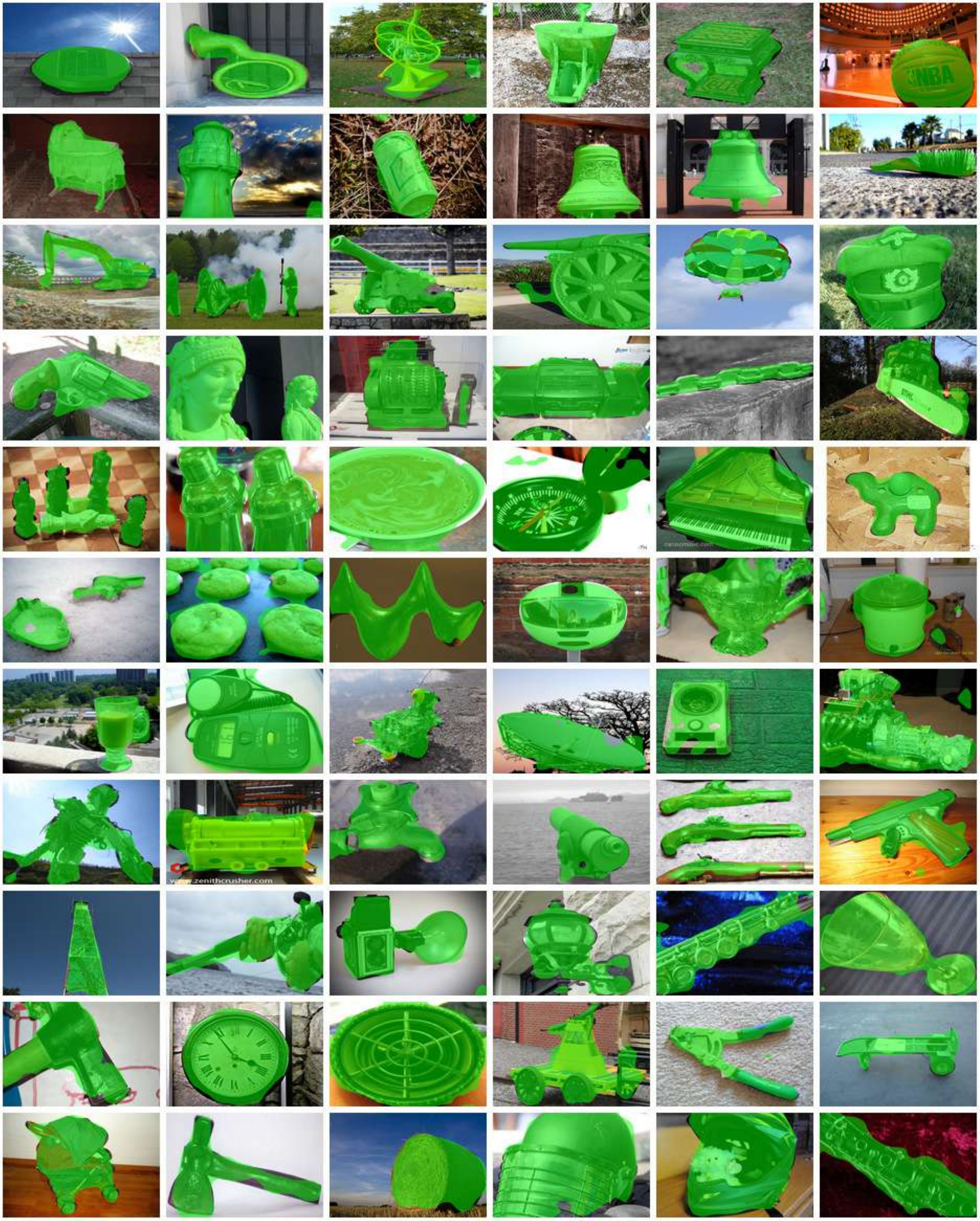} \\
	\hline
	\end{tabular}
	\caption{Qualitative results: We show example segmentations from ImageNet dataset obtained by our pixel objectness model on Non-PASCAL Categories. The segmentation results are shown with a green overlay. Our method generalizes remarkably well and is able to accurately segment foreground objects even for those categories which were never seen during training. Best viewed in color.}
	  \label{fig:qual_res_unseen2}
\end{figure*}

\begin{figure*}[h!]
  \centering
  \renewcommand{\tabcolsep}{0pt}
  \captionsetup{width=1\textwidth, font={footnotesize}, skip=2pt}
   \begin{tabular}{|c|}
      \hline
      {\bf Additional ImageNet Qualitative Examples from Non-PASCAL (unseen) Categories} \\
      \hline
    \includegraphics[width=0.9\columnwidth]{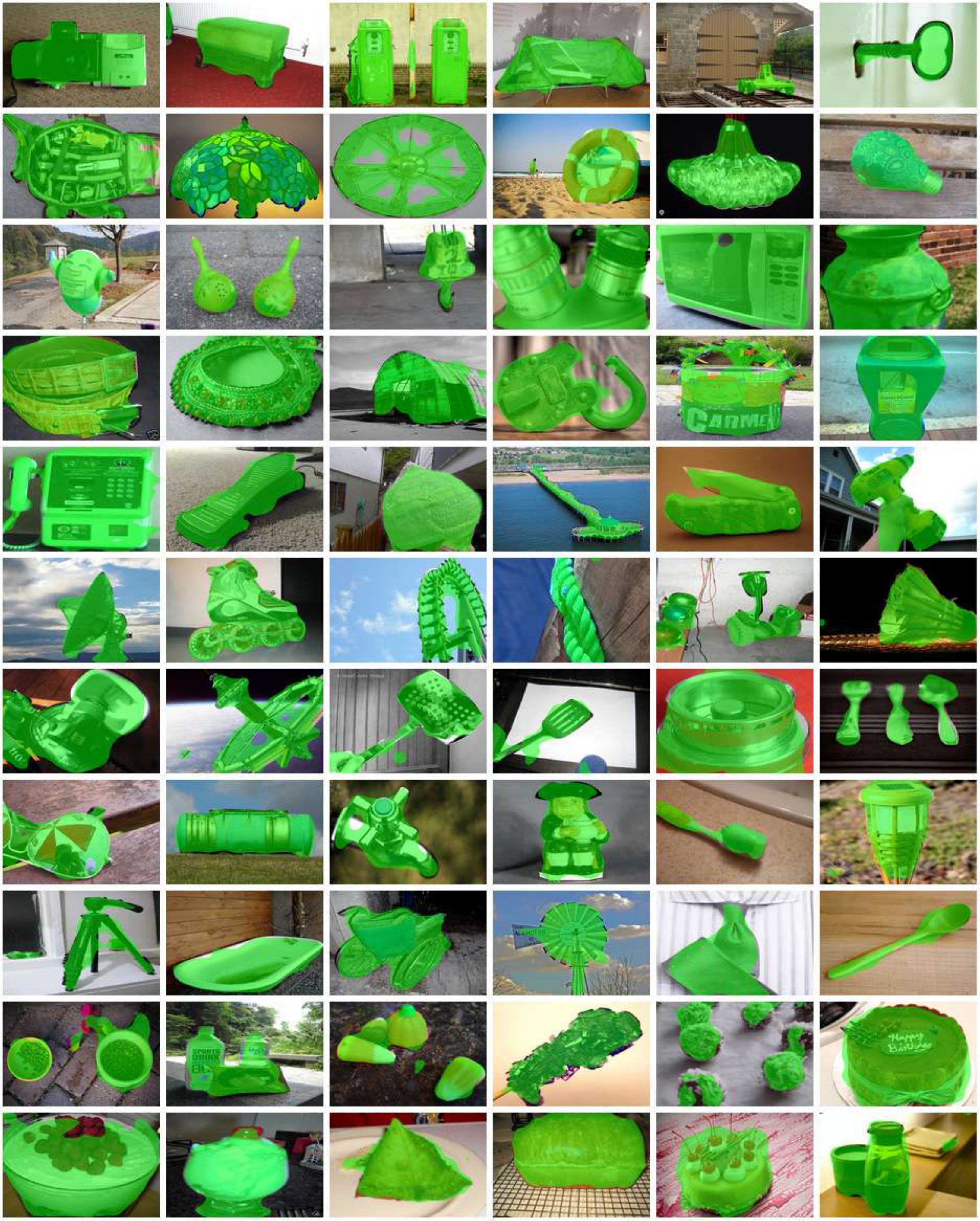} \\
	\hline
	\end{tabular}
	\caption{Qualitative results: We show example segmentations from ImageNet dataset obtained by our pixel objectness model on Non-PASCAL Categories. The segmentation results are shown with a green overlay. Our method generalizes remarkably well and is able to accurately segment foreground objects even for those categories which were never seen during training. Best viewed in color.}
	  \label{fig:qual_res_unseen3}
\end{figure*}

\begin{figure*}[h!]
  \centering
  \renewcommand{\tabcolsep}{0pt}
  \captionsetup{width=1\textwidth, font={footnotesize}, skip=2pt}
   \begin{tabular}{|c|}
      \hline
      {\bf Additional ImageNet Qualitative Examples for Failure Cases} \\
      \hline
    \includegraphics[width=0.9\columnwidth]{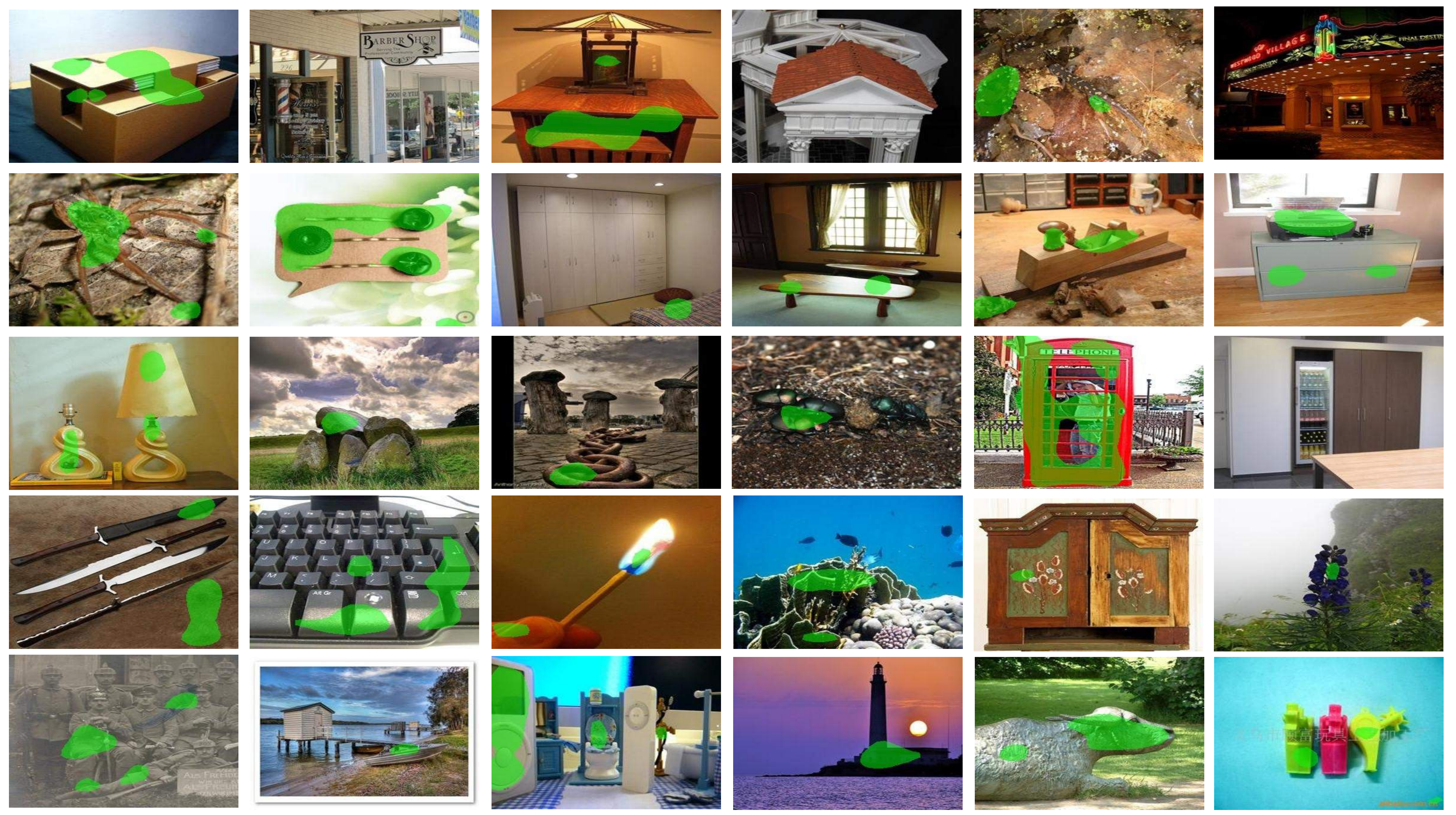} \\
        \hline
        \end{tabular}
        \caption{Qualitative results: We show examples of failure cases from ImageNet dataset obtained by our pixel objectness model. The segmentation results are shown with a green overlay. Typical failure cases involve scene-centric images or images containing very thin objects. Best viewed in color.}
          \label{fig:fail}
\end{figure*}

\clearpage

\subsection{Object-aware image retrieval analysis}

\begin{figure*}[t]
\centering
\includegraphics[width=1\columnwidth]{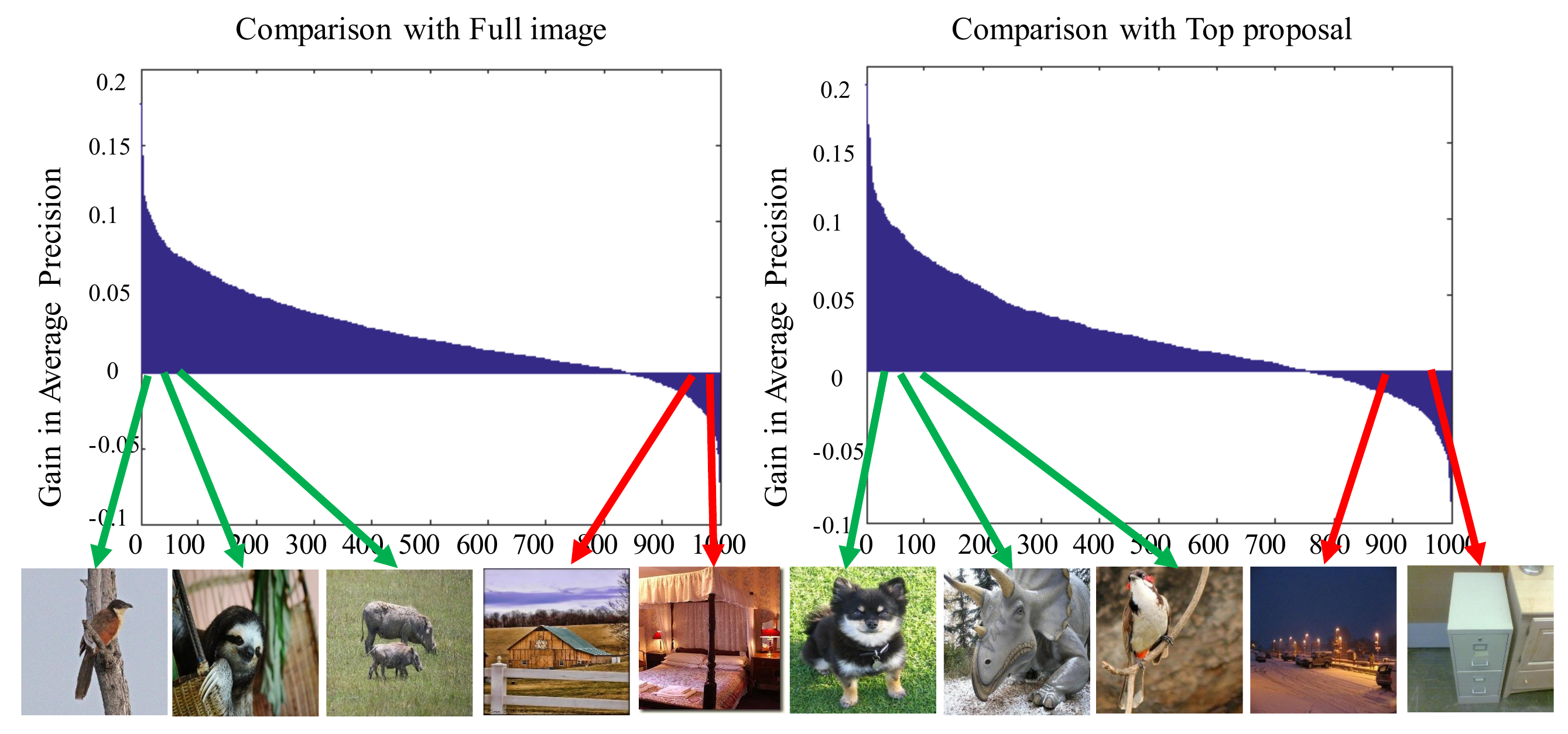}
\caption{
We show the \emph{gain} in average precision per object class between our method and the baselines (Full image on the left, and Top proposal on the right). Green arrows indicate example object classes for which our method performs better and red arrows indicate object classes for which the baselines perform better. Note our method excels at retrieving natural objects but can fail for scene-centric classes.}
\label{fig:category}
\end{figure*}

\begin{figure}[t]
\centering

\includegraphics[width=0.7\columnwidth]{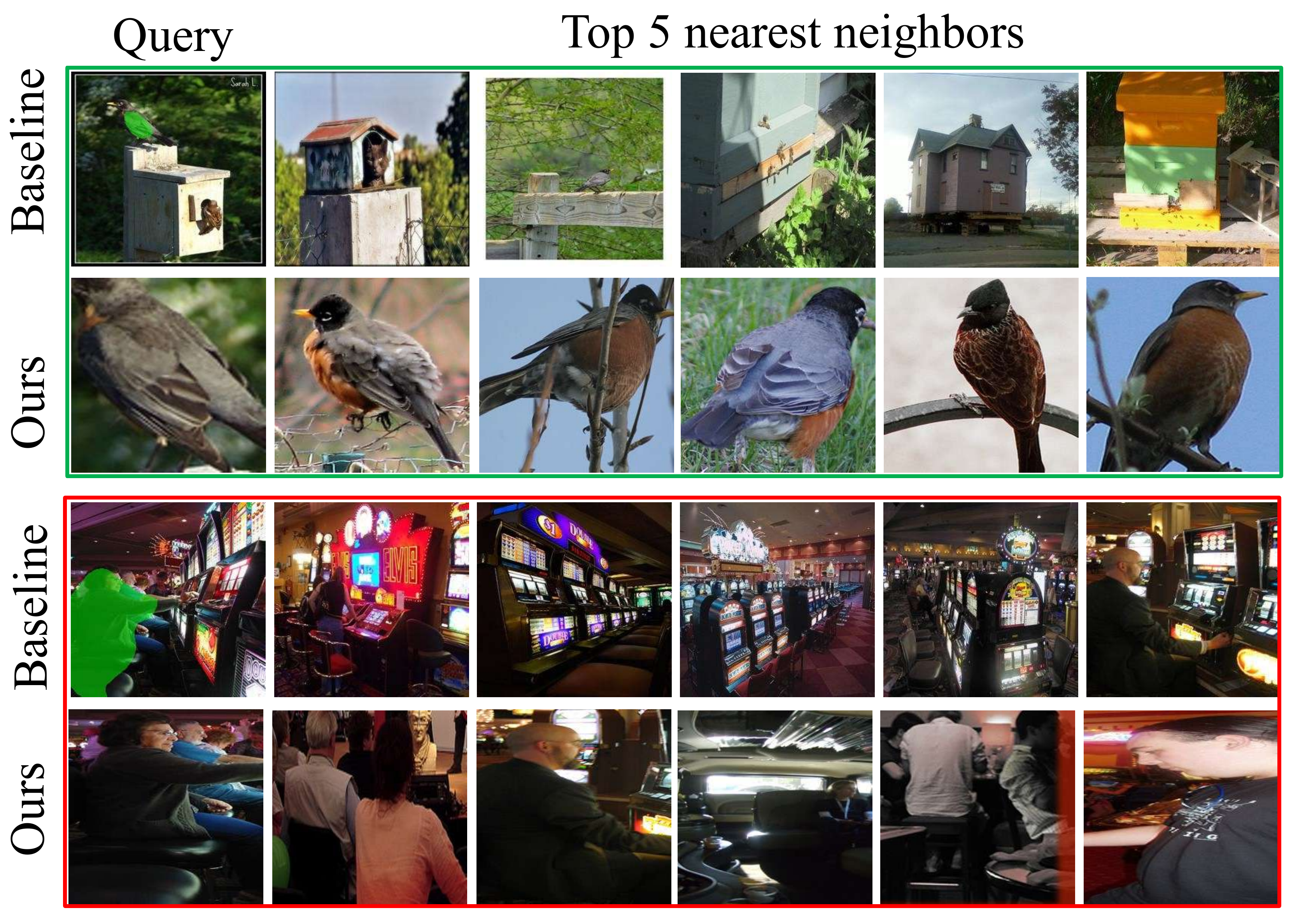}
\caption{Leveraging pixel objectness for object aware image retrieval (best viewed on pdf).}
\label{fig:nn}
\end{figure}

We next present more analysis for object-aware image retrieval. Please refer to \emph{Section~\ref{sec:results_pixel_app}} in the main paper for experiment details.
Figure~\ref{fig:category} looks more closely at the distribution of our method's gains in average precision per class. 
For the majority of classes, our method boosts results. We observe that our method performs extremely well on object-centric classes such as animals, but does not perform as well on some classes that are scene-centric, such as lakeshore or seashore.

To further understand the superior performance of our method for image retrieval, we show the Top-5 nearest neighbors for both our method and the Full image baseline in Figure~\ref{fig:nn}. In the first example (first and second rows), the query image contains a small bird. Our method is able to segment the bird and retrieves relevant images that also contain birds. The baseline, on the contrary, has noisier retrievals due to mixing the background. The last two rows show a case where, at least according to ImageNet labels, our method fails.   Our method segments the person, and then retrieves images containing a person from different scenes, whereas the baseline focuses on the entire image and retrieves similar scenes.

\subsection{Human study for foreground-aware image retargeting}

\bx{We provide more details on human study for foreground-aware image retargeting. Please refer to \emph{Section~\ref{sec:results_pixel_app}}. To quantify the results over all 500 images, we perform a human study on Amazon Mechanical Turk. 
Both methods are instructed to resize the source image to $2/3$ of its original size.
We present image pairs produced by our method and the baseline in arbitrary order and ask workers to rank which image is more likely to have been manipulated by a computer.  Each image pair is evaluated by three different workers. Workers found that $38.53\%$ of the time images produced by our method are more likely to have been manipulated by a computer, $48.87\%$ for the baseline; both methods tie $12.60\%$ of the time. Thus, human evaluation with non-experts demonstrates that our method outperforms the baseline. In addition, we also ask a vision expert familiar with image retargeting---but not involved in this project---to score the 500 image pairs 
with the same interface as the crowd workers.  The vision expert found our method performs better for $78\%$ of the images, baseline is better for $13\%$, and both methods tie for $9\%$ images. This further confirms that our foreground prediction can enhance image retargeting by defining a more semantically meaningful energy function.}

\subsection{Foreground-aware image retargeting examples}

\bx{We next present a larger figure of foreground-aware image retargeting examples (Fig.\emph{8} in main paper) in Figure~\ref{fig:seam_example2} and more foreground-aware image retargeting example results in Figure~\ref{fig:seam_example1}.} Please refer to \emph{Section~\ref{sec:results_pixel_app}} in the main paper for algorithmic details. Our method is able to preserve important objects in the images thanks to the proposed foreground segmentation method. The baseline produces images with important objects distorted, because gradient strength is not a good indicator for perceived content, especially when background is textured. We also present a few failure cases in the rightmost column. In the first example, our method is unsuccessful at predicting the skateboard as the foreground and therefore results in an image with skateboard distorted. In the second example, our method is able to detect and preserve all the people in the image. However, the background distortion creates artifacts that make the resulting image unpleasant to look at compared to the baseline. In the last example, our method misclassified the pillow as foreground and results in an image with an amplified pillow.

\begin{figure}[t]
  \centering

    \includegraphics[width=0.95\columnwidth]{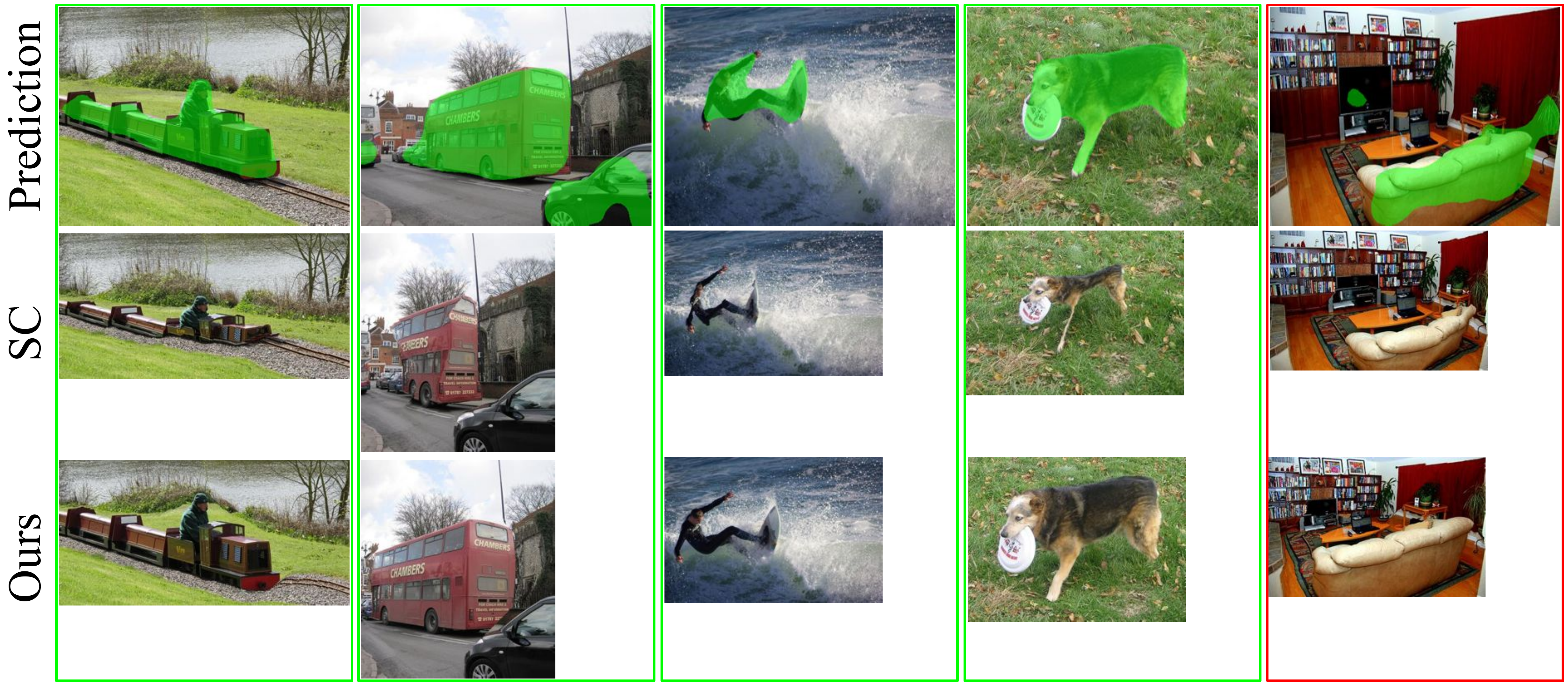} \\
   \caption{Leveraging pixel objectness for foreground aware image retargeting. Best viewed on pdf.}
\label{fig:seam_example2}
\end{figure}



\begin{figure}[h!]
\centering
\includegraphics[width=0.85\columnwidth]{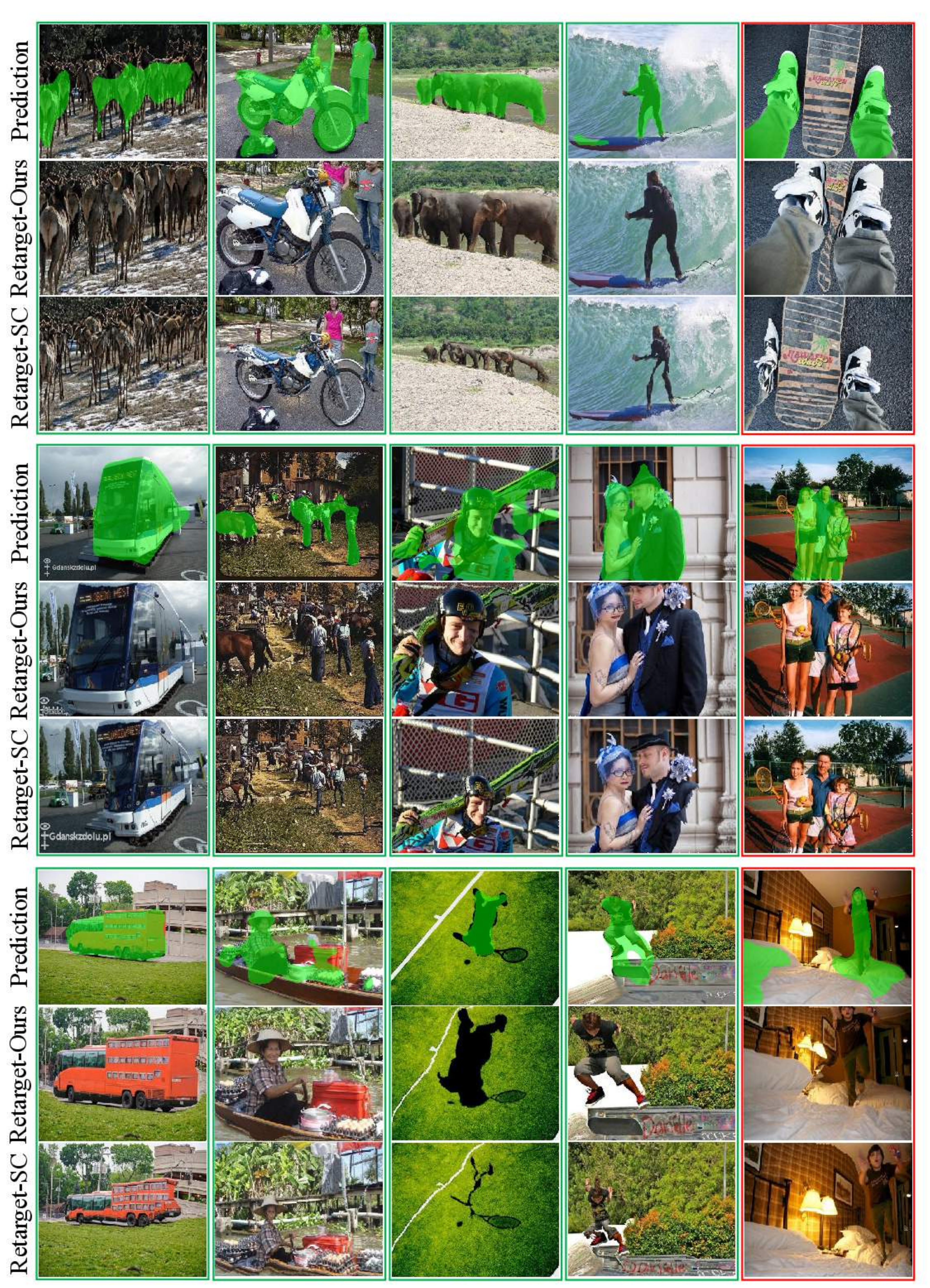}
\caption{We show more foreground-aware image retargeting example results. We show original images with predicted foreground in green (prediction, top row), retargeting images produced by our method (Retarget-Ours, middle row) and retargeting images produced by the Seam Carving based on gradient energy~\cite{avidan2007seam} (Retarget-SC, bottom row). Our method successfully preserves the important visual content while reducing the content of the background. We also present a few failure cases in the rightmost column. Best viewed in color.}
\label{fig:seam_example1}
\end{figure}
\clearpage

\begin{figure}[h!]
\centering
\includegraphics[width=0.95\columnwidth]{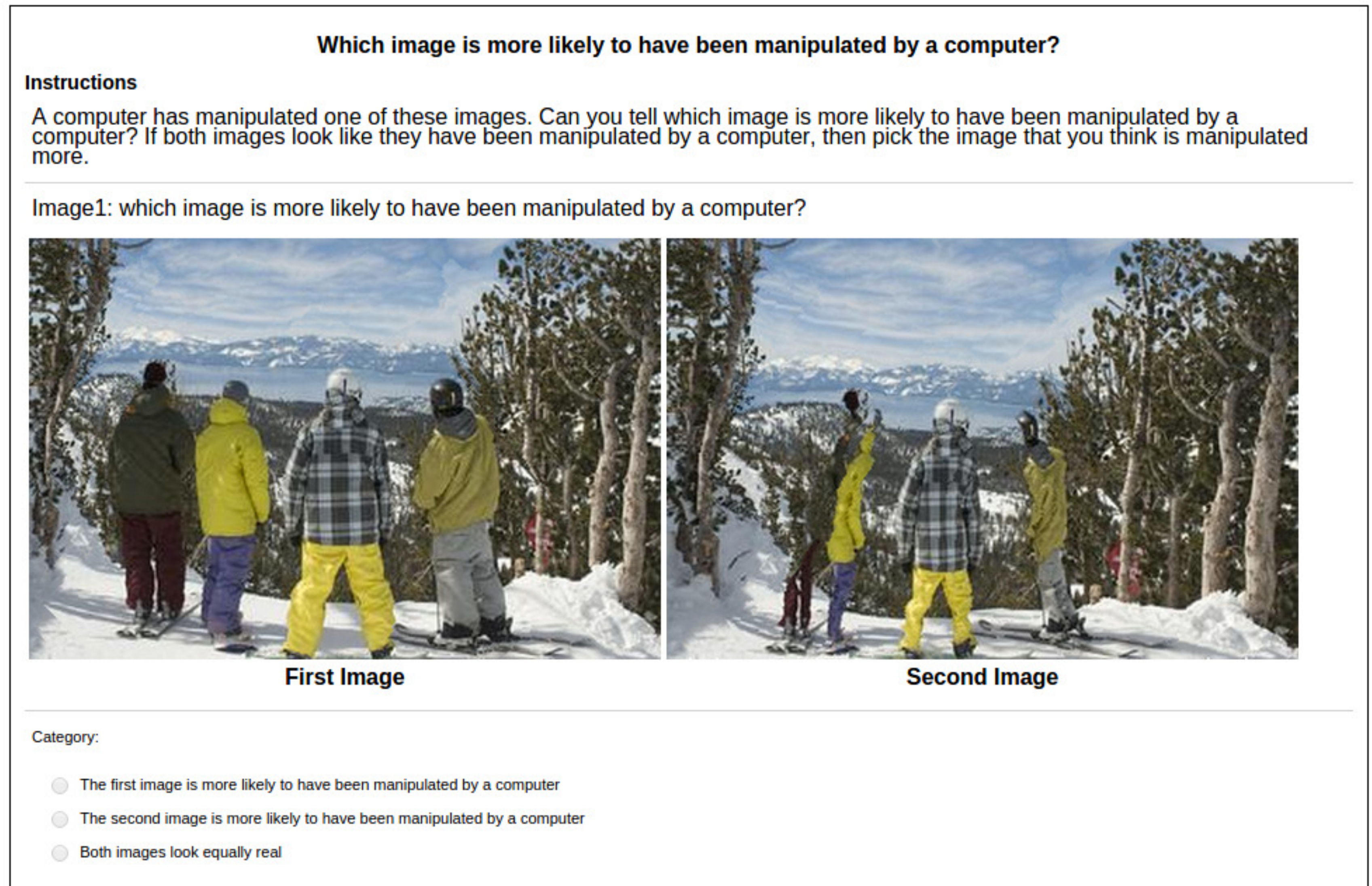}
\caption{Amazon Mechanical Turk interface used to collect human judgement for image retargeting. We ask workers to judge which image is more likely to have been manipulated by a computer. They have three options: 1) The first image is more likely to have been manipulated by a computer; 2) The second image is more likely to have been manipulated by a computer; 3) Both images look equally real.
}
\label{fig:amt_interface}
\end{figure}

\subsection{Amazon Mechanical Turk interface}

We also show the interface we used to collect human judgement for image retargeting on Amazon Mechanical Turk. The two images produced by our algorithm and the baseline method are shown in arbitrary order to the workers. We instruct the workers to pick an image that is more likely to have been manipulated by a computer. If both images look like they have been manipulated by a computer, then pick the one that is manipulated more. The workers have three options to choose from: 1) The first image is more likely to have been manipulated by a computer; 2) The second image is more likely to have been manipulated by a computer; 3) Both images look equally real. See Figure~\ref{fig:amt_interface} for the interface. Also refer to \emph{Section~\ref{sec:results_pixel_app}} in the main paper for more discussions on these user study results.

\section{Per-video results for DAVIS and Segtrack-v2} Table~\ref{davis-results_s} shows the per video results for the 50 videos from the DAVIS dataset (referenced in Table 4 of the main paper). We compare with several semi-supervised and fully automatic baselines. 

Table~\ref{segtrack-results_s} shows the per video results for the 14 videos from the Segtrack-v2 dataset (referenced in Table 6 of the main paper). Our method outperforms the per-video best fully automatic method in 5 out of 14 cases. 

\begin{table*}[h!]
	\centering
	\footnotesize
	\begin{tabular}{|c|ccc|ccc|ccc|}
		\hline
		\multicolumn{10}{|c|}{DAVIS: Densely Annotated Video Segmentation dataset (50 videos)}  \\
		\hline
		\hline
		Methods & FST~\cite{ferrari-iccv2013} & KEY~\cite{keysegments} & NLC~\cite{nlc} & HVS~\cite{grundmann-cvpr2010} & FCP~\cite{Perazzi_2015_ICCV} & BVS~\cite{marki2016bilateral} & Ours-A & Ours-M & Ours-Joint \\
		
		\hline 
        Human in loop?  & No & No & No & Yes & Yes & Yes & No & No & No \\       
		\hline
		\hline		
		Bear & 89.8 & 89.1 & 90.7 & 93.8 & 90.6 & 95.5 & 91.52 & 86.30 & 91.72 \\
		Blackswan & 73.2 & 84.2 & 87.5 & 91.6 & 90.8 & 94.3 & 89.54 & 61.71 & 87.20 \\
		Bmx-Bumps & 24.1 & 30.9 & 63.5 & 42.8 & 30 & 43.4 & 38.77 & 26.42 & 40.94 \\
		Bmx-Trees & 18 & 19.3 & 21.2 & 17.9 & 24.8 & 38.2 & 34.67 & 37.08 & 52.77 \\
		Boat & 36.1 & 6.5 & 0.7 & 78.2 & 61.3 & 64.4 & 63.80 & 59.53 & 70.63 \\
		Breakdance & 46.7 & 54.9 & 67.3 & 55 & 56.7 & 50 & 14.22 & 61.80 & 31.37 \\
		Breakdance-Flare & 61.6 & 55.9 & 80.4 & 49.9 & 72.3 & 72.7 & 54.87 & 62.09 & 80.67 \\
		Bus & 82.5 & 78.5 & 62.9 & 80.9 & 83.2 & 86.3 & 80.38 & 77.70 & 83.90 \\
		Camel & 56.2 & 57.9 & 76.8 & 87.6 & 73.4 & 66.9 & 76.39 & 74.19 & 78.86 \\
		Car-Roundabout & 80.8 & 64 & 50.9 & 77.7 & 71.7 & 85.1 & 74.84 & 84.75 & 81.34 \\
		Car-Shadow & 69.8 & 58.9 & 64.5 & 69.9 & 72.3 & 57.8 & 88.38 & 81.03 & 92.43 \\
		Car-Turn & 85.1 & 80.6 & 83.3 & 81 & 72.4 & 84.4 & 90.67 & 83.92 & 91.81 \\
		Cows & 79.1 & 33.7 & 88.3 & 77.9 & 81.2 & 89.5 & 87.96 & 82.22 & 89.13 \\
		Dance-Jump & 59.8 & 74.8 & 71.8 & 68 & 52.2 & 74.5 & 10.32 & 64.22 & 48.60 \\
		Dance-Twirl & 45.3 & 38 & 34.7 & 31.8 & 47.1 & 49.2 & 46.23 & 55.39 & 75.76 \\
		Dog & 70.8 & 69.2 & 80.9 & 72.2 & 77.4 & 72.3 & 90.41 & 81.90 & 91.38 \\
		Dog-Agility & 28 & 13.2 & 65.2 & 45.7 & 45.3 & 34.5 & 68.94 & 67.88 & 74.52 \\
		Drift-Chicane & 66.7 & 18.8 & 32.4 & 33.1 & 45.7 & 3.3 & 46.13 & 44.14 & 73.27 \\
		Drift-Straight & 68.3 & 19.4 & 47.3 & 29.5 & 66.8 & 40.2 & 67.24 & 69.08 & 83.04 \\
		Drift-Turn & 53.3 & 25.5 & 15.4 & 27.6 & 60.6 & 29.9 & 85.09 & 72.09 & 89.69 \\
		Elephant & 82.4 & 67.5 & 51.8 & 74.2 & 65.5 & 85 & 86.18 & 77.51 & 88.43 \\
		Flamingo & 81.7 & 69.2 & 53.9 & 81.1 & 71.7 & 88.1 & 44.46 & 63.80 & 56.98 \\
		Goat & 55.4 & 70.5 & 1 & 58 & 67.7 & 66.1 & 84.11 & 74.99 & 83.23 \\
		Hike & 88.9 & 89.5 & 91.8 & 87.7 & 87.4 & 75.5 & 82.54 & 58.30 & 81.86 \\
		Hockey & 46.7 & 51.5 & 81 & 69.8 & 64.7 & 82.9 & 66.03 & 44.89 & 72.00 \\
		Horsejump-High & 57.8 & 37 & 83.4 & 76.5 & 67.6 & 80.1 & 71.09 & 54.10 & 75.16 \\
		Horsejump-Low & 52.6 & 63 & 65.1 & 55.1 & 60.7 & 60.1 & 70.23 & 55.20 & 77.68 \\
		Kite-Surf & 27.2 & 58.5 & 45.3 & 40.5 & 57.7 & 42.5 & 47.71 & 18.54 & 51.58 \\
		Kite-Walk & 64.9 & 19.7 & 81.3 & 76.5 & 68.2 & 87 & 52.65 & 39.35 & 52.45 \\
		Libby & 50.7 & 61.1 & 63.5 & 55.3 & 31.6 & 77.6 & 67.70 & 35.34 & 61.29 \\
		Lucia & 64.4 & 84.7 & 87.6 & 77.6 & 80.1 & 90.1 & 79.93 & 49.18 & 79.07 \\
		Mallard-Fly & 60.1 & 58.5 & 61.7 & 43.6 & 54.1 & 60.6 & 74.62 & 42.64 & 74.23 \\
		Mallard-Water & 8.7 & 78.5 & 76.1 & 70.4 & 68.7 & 90.7 & 83.34 & 25.31 & 82.91 \\
		Motocross-Bumps & 61.7 & 68.9 & 61.4 & 53.4 & 30.6 & 40.1 & 83.78 & 56.56 & 82.04 \\
		Motocross-Jump & 60.2 & 28.8 & 25.1 & 9.9 & 51.1 & 34.1 & 80.43 & 59.02 & 80.43\\
		Motorbike & 55.9 & 57.2 & 71.4 & 68.7 & 71.3 & 56.3 & 28.67 & 45.71 & 40.45 \\
		Paragliding & 72.5 & 86.1 & 88 & 90.7 & 86.6 & 87.5 & 17.68 & 60.76 & 62.23 \\
		Paragliding-Launch & 50.6 & 55.9 & 62.8 & 53.7 & 57.1 & 64 & 58.88 & 50.34 & 41.61 \\
		Parkour & 45.8 & 41 & 90.1 & 24 & 32.2 & 75.6 & 79.39 & 58.51 & 78.06 \\
		Rhino & 77.6 & 67.5 & 68.2 & 81.2 & 79.4 & 78.2 & 77.56 & 83.03 & 88.22 \\
		Rollerblade & 31.8 & 51 & 81.4 & 46.1 & 45 & 58.8 & 63.27 & 57.73 & 70.05 \\
		Scooter-Black & 52.2 & 50.2 & 16.2 & 62.4 & 50.4 & 33.7 & 36.07 & 62.18 & 68.05 \\
		Scooter-Gray & 32.5 & 36.3 & 58.7 & 43.3 & 48.3 & 50.8 & 73.22 & 61.69 & 79.23 \\
		Soapbox & 41 & 75.7 & 63.4 & 68.4 & 44.9 & 78.9 & 49.70 & 53.24 & 71.29 \\
		Soccerball & 84.3 & 87.9 & 82.9 & 6.5 & 82 & 84.4 & 29.27 & 73.56 & 46.96 \\
		Stroller & 58 & 75.9 & 84.9 & 66.2 & 59.7 & 76.7 & 63.91 & 54.40 & 70.14 \\
		Surf & 47.5 & 89.3 & 77.5 & 75.9 & 84.3 & 49.2 & 88.78 & 73.00 & 89.88 \\
		Swing & 43.1 & 71 & 85.1 & 10.4 & 64.8 & 78.4 & 73.75 & 59.41 & 77.26 \\
		Tennis & 38.8 & 76.2 & 87.1 & 57.6 & 62.3 & 73.7 & 76.88 & 47.19 & 77.98 \\
		Train & 83.1 & 45 & 72.9 & 84.6 & 84.1 & 87.2 & 42.50 & 80.33 & 71.45 \\
		\hline
		\hline
		Avg. IoU  & 57.5 & 56.9 & 64.1 & 59.6 & 63.1 & {\bf 66.5} & 64.69 & 60.18 & {\bf 72.82} \\
		\hline
	\end{tabular}
	\caption{Video object segmentation results on DAVIS dataset. We show the results for all 50 videos. Table 4 in the main paper summarizes these results over all 50 videos. Our method outperforms several state-of-the art methods, including the ones which actually require human annotation during segmentation. The best performing methods grouped by whether they require human-in-the-loop or not during segmentation are highlighted in bold. Metric: Jaccard score, higher is better. }
	\label{davis-results_s}
\end{table*}

\begin{table*}[t]
	\centering
	\begin{tabular}{|c|ccc|c|ccc|}
		\hline
		\multicolumn{8}{|c|}{Segtrack-v2 dataset (14 videos)}  \\
		\hline		
		\hline		
		Methods  & FST~\cite{ferrari-iccv2013} & KEY~\cite{keysegments} & NLC~\cite{nlc}  & HVS~\cite{grundmann-cvpr2010} & Ours-A & Ours-M & Ours-Joint \\
		
		\hline	
		Human in loop?  & No & No & No & Yes  & No & No & No \\
		\hline	
		\hline
		birdfall2        & 17.50 & 49.00 & 74.00 & 57.40 & 6.94  & 55.50 & 16.03 \\
		bird of paradise & 81.83 & 92.20 & -     & 86.80 & 49.82 & 62.46 & 65.63 \\
		bmx              & 67.00 & 63.00 & 79.00 & 35.85 & 59.53 & 55.12 & 61.50 \\
		cheetah          & 28.00 & 28.10 & 69.00 & 21.60 & 71.15 & 36.00 & 65.39 \\
		drift            & 60.50 & 46.90 & 86.00 & 41.20 & 82.18 & 80.03 & 88.75 \\
		frog             & 54.13 & 0.00  & 83.00 & 67.10 & 54.86 & 52.88 & 63.52 \\
		girl             & 54.90 & 87.70 & 91.00 & 31.90 & 81.07 & 43.57 & 77.07 \\
		hummingbird      & 52.00 & 60.15 & 75.00 & 19.45 & 61.50 & 60.86 & 69.96 \\
		monkey           & 65.00 & 79.00 & 71.00 & 61.90 & 86.42 & 58.95 & 86.04 \\
		monkeydog        & 61.70 & 39.60 & 78.00 & 43.55 & 39.08 & 24.36 & 42.26 \\
		parachute        & 76.32 & 96.30 & 94.00 & 69.10 & 24.86 & 59.43 & 74.48 \\
		penguin          & 18.31 & 9.27  & -     & 74.45 & 66.20 & 45.09 & 61.42 \\
		soldier          & 39.77 & 66.60 & 83.00 & 66.50 & 83.70 & 48.37 & 76.19 \\
		worm             & 72.79 & 84.40 & 81.00 & 34.70 & 29.13 & 59.94 & 53.96 \\
		\hline 
		Avg. IoU  & 53.5 & 57.3 & {\bf 80\textsuperscript{*} }  & {\bf 50.8} & 56.88 & 53.04 & 64.44 \\
		\hline	
	\end{tabular}
	\caption{Video object segmentation results on Segtrack-v2. We show the results for all 14 videos. Table 6 in the main paper summarizes these results over all 14 videos. Our method outperforms several state-of-the art methods, including the ones which actually require human annotation during segmentation. For NLC results are averaged over 12 videos as reported in their paper~\cite{nlc}. The best performing methods grouped by whether they require human-in-the-loop or not during segmentation are highlighted in bold.  Metric: Jaccard score, higher is better. }
	\label{segtrack-results_s}
	\vspace{-8pt}
\end{table*}







\end{document}